\documentclass[]{IEEEtran}
\ifCLASSINFOpdf
\else
\fi
\usepackage{amsmath,amssymb,amsfonts}
\usepackage{algorithmic}
\usepackage[ruled,vlined,linesnumbered]{algorithm2e}
\usepackage{algorithmic,algorithm2e,float}
\SetKwInput{KwInput}{Input}                % Set the Input
\SetKwInput{KwOutput}{Output}              % set the Output
\SetKwInput{KwRepeat}{Repeat}
\SetKwInput{KwUntil}{Until}
\SetKwInput{KwReturn}{Return}
\usepackage{amsfonts}
\usepackage{graphicx}
\usepackage{textcomp}
\usepackage{xcolor}
\usepackage{tabularx}
\usepackage{array}
\usepackage{booktabs}
\usepackage{threeparttable}  
\usepackage{multirow}
\usepackage{bm}
\usepackage{amsmath}
\usepackage{subfigure}
\usepackage{amssymb}% http://ctan.org/pkg/amssymb
\usepackage{pifont}% http://ctan.org/pkg/pifont
\newcommand{\cmark}{\ding{51}}%
\newcommand\drop[1]{{\color{red}#1}}

\begin{document}
%
% paper title
% Titles are generally capitalized except for words such as a, an, and, as,
% at, but, by, for, in, nor, of, on, or, the, to and up, which are usually
% not capitalized unless they are the first or last word of the title.
% Linebreaks \\ can be used within to get better formatting as desired.
% Do not put math or special symbols in the title.
\title{LGGNet: Learning from Local-Global-Graph Representations for Brain-Computer Interface}
%
%
% author names and IEEE memberships
% note positions of commas and nonbreaking spaces ( ~ ) LaTeX will not break
% a structure at a ~ so this keeps an author's name from being broken across
% two lines.
% use \thanks{} to gain access to the first footnote area
% a separate \thanks must be used for each paragraph as LaTeX2e's \thanks
% was not built to handle multiple paragraphs
%

\author{Yi~Ding,~\IEEEmembership{Student Member,~IEEE,}
        Neethu~Robinson,~\IEEEmembership{Member,~IEEE,}
        Chengxuan~Tong,~\IEEEmembership{Student Member,~IEEE,}
        Qiuhao~Zeng,
        and~Cuntai~Guan,~\IEEEmembership{Fellow,~IEEE}% <-this % stops a space
\thanks{Yi Ding, Neethu Robinson, Chengxuan Tong, Qiuhao Zeng, and Cuntai Guan are with the School of Computer Science and Engineering, Nanyang Technological University, 50 Nanyang Avenue, Singapore, 639798. e-mail: (ding.yi, nrobinson, tong0110, qiuhao.zeng, ctguan)@ntu.edu.sg.}% <-this % stops a space
% <-this % stops a space
\thanks{Cuntai Guan is the Corresponding Author.}
}
%\thanks{Manuscript received April %19, 2005; revised August 26, 2015.}}

% note the % following the last \IEEEmembership and also \thanks - 
% these prevent an unwanted space from occurring between the last author name
% and the end of the author line. i.e., if you had this:
% 
% \author{....lastname \thanks{...} \thanks{...} }
%                     ^------------^------------^----Do not want these spaces!
%
% a space would be appended to the last name and could cause every name on that
% line to be shifted left slightly. This is one of those "LaTeX things". For
% instance, "\textbf{A} \textbf{B}" will typeset as "A B" not "AB". To get
% "AB" then you have to do: "\textbf{A}\textbf{B}"
% \thanks is no different in this regard, so shield the last } of each \thanks
% that ends a line with a % and do not let a space in before the next \thanks.
% Spaces after \IEEEmembership other than the last one are OK (and needed) as
% you are supposed to have spaces between the names. For what it is worth,
% this is a minor point as most people would not even notice if the said evil
% space somehow managed to creep in.

% The paper headers
\markboth{IEEE Transactions on Neural Networks and Learning Systems}%
{Shell \MakeLowercase{\textit{et al.}}: Bare Demo of IEEEtran.cls for IEEE Journals}
% The only time the second header will appear is for the odd numbered pages
% after the title page when using the twoside option.
% 
% *** Note that you probably will NOT want to include the author's ***
% *** name in the headers of peer review papers.                   ***
% You can use \ifCLASSOPTIONpeerreview for conditional compilation here if
% you desire.

% If you want to put a publisher's ID mark on the page you can do it like
% this:
%\IEEEpubid{0000--0000/00\$00.00~\copyright~2015 IEEE}
% Remember, if you use this you must call \IEEEpubidadjcol in the second
% column for its text to clear the IEEEpubid mark.

% use for special paper notices
%\IEEEspecialpapernotice{(Invited Paper)}

% make the title area
\maketitle

% As a general rule, do not put math, special symbols or citations
% in the abstract or keywords.
\begin{abstract}
Neuropsychological studies suggest that co-operative activities among different brain functional areas drive high-level cognitive processes. To learn the brain activities within and among different functional areas of the brain, we propose LGGNet, a novel neurologically inspired graph neural network, to learn local-global-graph representations of electroencephalography (EEG) for Brain-Computer Interface (BCI). The input layer of LGGNet comprises a series of temporal convolutions with multi-scale 1D convolutional kernels and kernel-level attentive fusion. It captures temporal dynamics of EEG which then serves as input to the proposed local and global graph-filtering layers. Using a defined neurophysiologically meaningful set of local and global graphs, LGGNet models the complex relations within and among functional areas of the brain. Under the robust nested cross-validation settings, the proposed method is evaluated on three publicly available datasets for four types of cognitive classification tasks, namely, the attention, fatigue, emotion, and preference classification tasks. LGGNet is compared with state-of-the-art methods, such as DeepConvNet, EEGNet, R2G-STNN, TSception, RGNN, AMCNN-DGCN, HRNN and GraphNet. The results show that LGGNet outperforms these methods, and the improvements are statistically significant ($p<0.05$) in most cases. The results show that bringing neuroscience prior knowledge into neural network design yields an improvement of classification performance. The source code can be found at \textit{https://github.com/yi-ding-cs/LGG}
\end{abstract}

% Note that keywords are not normally used for peerreview papers.

% For peer review papers, you can put extra information on the cover
% page as needed:
% \ifCLASSOPTIONpeerreview
% \begin{center} \bfseries EDICS Category: 3-BBND \end{center}
% \fi
%
% For peerreview papers, this IEEEtran command inserts a page break and
% creates the second title. It will be ignored for other modes.
\IEEEpeerreviewmaketitle

\section{Introduction}
\IEEEPARstart{B}{rain}-computer interface (BCI) enables the brain to communicate with machines directly using electroencephalography (EEG) \cite{9093122}. A typical BCI system consists of a data acquisition module, a pre-processing module, a classification module, and a feedback module \cite{lotte2010regularizing}. BCI has a wide range of applications in the real world, such as robot controlling \cite{liu2015fdes}, stroke rehabilitation \cite{foong2019assessment}, and emotion regulation for mental disorders \cite{zotev2020emotion, XING2019338}.

Compared with traditional machine learning methods \cite{4634130, 7078926, 7938737, 8634938}, deep learning methods achieved superior performances in different tasks of BCI, such as classification of motor imagery \cite{doi:10.1002/hbm.23730,8897723,Tabar_2016,Lawhern_2018}, mental attention classification \cite{Fahimi_2019,9401744,9361688}, emotion recognition \cite{7946165, Li2018, 8567966, 9206750}, and mental workload detection \cite{JIAO2018582}. However, most of the previous studies highly rely on manually extracted EEG features, such as power spectral density (PSD) \cite{9361688, JIAO2018582} and differential entropy (DE) \cite{Li2018, 8320798, 9091308}. With the feature-extracting ability of convolutional neural networks (CNNs), directly learning from EEG becomes reliable \cite{doi:10.1002/hbm.23730, Lawhern_2018, 9206750}. There are mainly two types of information to be learned in EEG, temporal and spatial information. The temporal information is well studied by the 1D CNNs \cite{doi:10.1002/hbm.23730, Lawhern_2018} and multi-scale 1D CNNs \cite{9206750}. For spatial information, previous methods either learn global spatial information using 1D CNNs along electrode dimension \cite{doi:10.1002/hbm.23730, Lawhern_2018, 9206750} or apply small 2D CNN kernels on image-liked EEG 2D maps \cite{Li2018, JIAO2018582, 8713896} to extract local spatial information separately, which may not learn the spatial information effectively. EEG signals can be naturally regarded as graph-structured data, with each electrode being the node and spatial relations \cite{9091308} or correlations among electrodes \cite{8462207} being the edges. A graph neural network (GNN) with proper adjacency relations can jointly learn the localized and global spatial patterns in EEG.

Incorporating the prior knowledge from neuropsychological studies into GNN design has huge potentials in mental states decoding from EEG. The brain is a complex network with a hierarchical spatial and functional organization at the level of neurons, local circuits, and functional areas \cite{POWER2011665}. Different functional areas correlate to certain brain functions while not working independently \cite{Friederici2017}. Activating one particular brain region also tends to activate other regions in the group \cite{KOBER2008998}. How to design neurophysiologically meaningful networks to effectively model the brain activities within and among different functional areas of the brain becomes crucial. Some studies \cite{8320798, 9309090} used a global adjacency matrix with learnable connections which paid less attention to the localized activities in each functional area. RGNN \cite{9091308} built the connections according to the spatial distance among electrodes. Although it added fixed global connections to improve the decoding performance, the complex relations among functional areas were not learned capably. 

To address the above problems, we propose to define the EEG data as a local-global graph whose local graphs belong to the different functional areas of the brain according to neurological knowledge \cite{POWER2011665, doi:10.1111/psyp.13028}. The nodes in each local graph are fully connected because they reflect the brain activities within each brain functional area. The edges of local graphs, or the global connections among local graphs, reflect the complex functional connections among different brain functional regions. To extract more information-rich representations from EEG as the node attributes in the proposed local-global-graph representations of EEG, a temporal convolutional layer with multi-scale 1D convolutional kernels is adopted \cite{9206750}. A kernel-level attentive fusion layer is further designed to fuse the learned temporal representations with attention. For graph connection learning, a local graph-filtering layer and a global graph-filtering layer are proposed to learn the brain activities within and among different local graphs. In the local graph-filtering layer, the attributes of the nodes are attentively aggregated into one hidden embedding which represents the activity of the local graph. For the global graph-filtering layer, an instance-specific similarity matrix is proposed as the base adjacency matrix of the global graph. Inspired by DGCNN \cite{8320798}, a learnable adjacency mask is further utilized to select the global connections attentively via back-propagation during the training process. We propose general, frontal, and hemisphere local-global-graph definitions of EEG based on neurophysiological evidence of associations among brain areas for different mental tasks. The general local-global graph is defined according to the 10-20 system that groups the electrode based on the location of electrodes on functional areas \cite{7946165}. In the frontal local-global graph, the frontal region is further divided into smaller local regions which are symmetrically located on the left and right hemispheres to learn asymmetric patterns in emotion \cite{doi:10.1111/psyp.13028}. In the hemisphere local-global graph, the symmetrically located sub-graphs exist in all the functional areas. 

In this paper, we propose Local-Global-Graph Network (LGGNet) that integrates all the aforementioned learning blocks to model the activities within and among brain functional areas for mental state classification. LGGNet was evaluated on four classification tasks, attention, fatigue, emotion, and preference classification, using three publicly available benchmark datasets, the attention dataset \cite{Shin2018}, the fatigue dataset \cite{cao2019multi}, and the DEAP dataset \cite{5871728}, respectively. The proposed LGGNet was compared with several state-of-the-art (SOTA) methods in the BCI domain. From the experiment results, LGGNet achieved the highest accuracies and F1 scores among the compared SOTA methods in most of the classification experiments. Furthermore, ablation studies were conducted to understand the importance of kernel-level attentive fusion, local and global graph-filtering layers in LGGNet. To evaluate the effectiveness of involving neuroscientific prior knowledge in LGGNet, the effect of building EEG as local-global graphs as well as the differences among different graph definitions were analyzed. After that, extensive visualization experiments were conducted to better understand what the network learns from EEG. The most informative region of the data identified by the network was visualized using saliency maps \cite{simonyan2013deep}. The learned adjacency matrices for different cognitive tasks were visualized as well. 

The major contributions of this work can be summarised as: 
\begin{itemize}
\item Proposed LGGNet, a neurologically inspired graph neural network, to learn the brain activities within and among different brain functional areas. 

\item Three different types of local-global graphs, namely the general, frontal, and hemisphere local-global graphs, were proposed to study the effects of different graph definitions on different cognitive tasks. 

\item The proposed method was compared with DeepConvNet (2017) \cite{doi:10.1002/hbm.23730}, EEGNet (2018) \cite{Lawhern_2018} , R2G-STNN (2019) \cite{8736804},  TSception (2020) \cite{9206750}, RGNN (2020) \cite{9091308}, AMCNN-DGCN (2021) \cite{9309090}, HRNN (2021) \cite{9361688}, GraphNet (2021) \cite{9361688} on three publicly available datasets for four different types of cognitive tasks: attention, fatigue, emotion, and preference classification. 

\item Extensive ablation studies and analysis experiments were conducted to better understand LGGNet.
\end{itemize}

The remainder of this article is organized as follows. Some related work is given in Section II. In Section III, the proposed LGGNet is introduced. In Section IV, the dataset and experiment settings are presented. The result and discussion are provided in Section V. Finally, we conclude the paper in Section VI.
\begin{figure*}[ht]
    \centering
    \includegraphics[width= \linewidth]{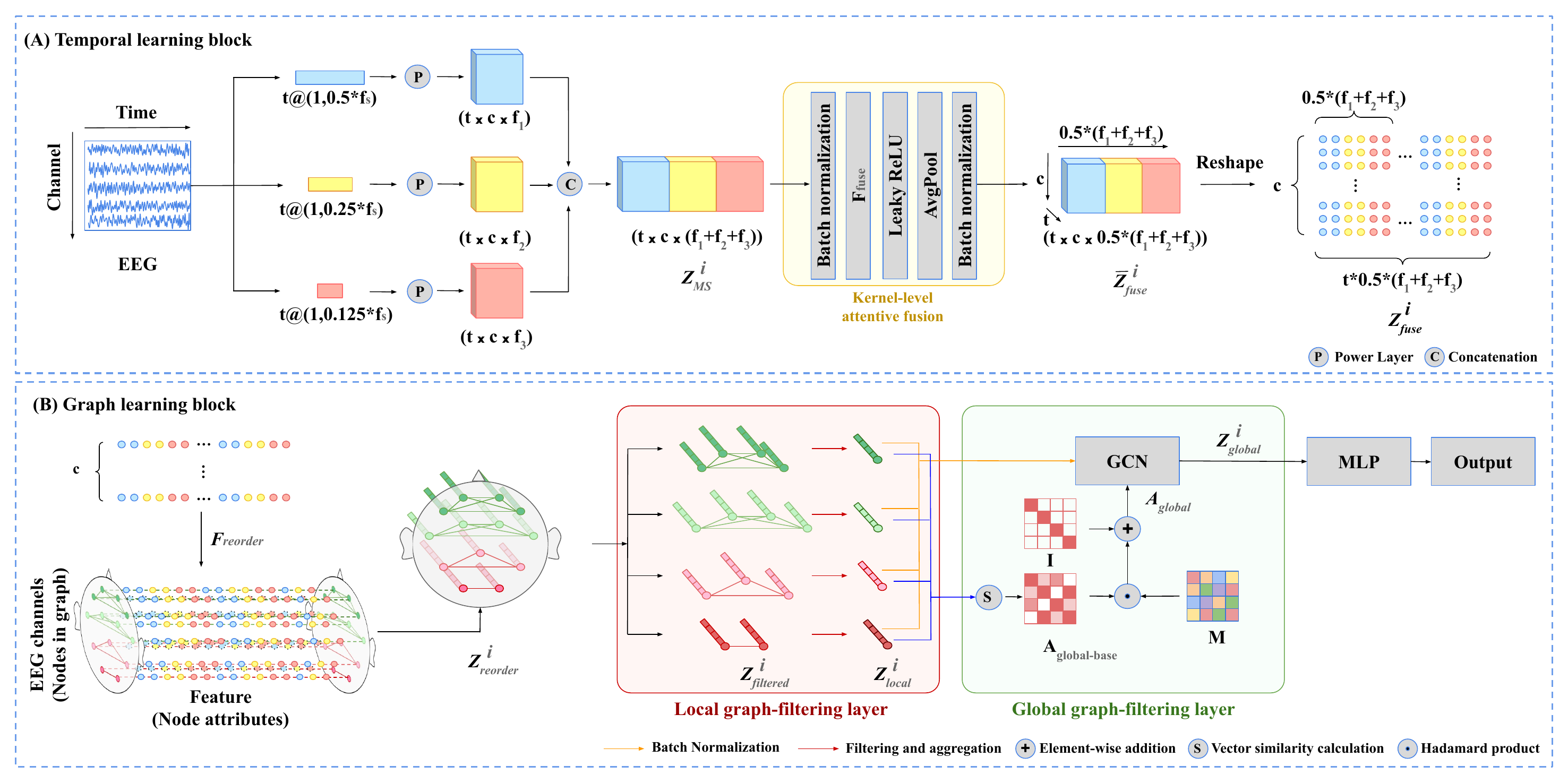}
    \caption{Structure of LGGNet. LGGNet has two main functional blocks: the temporal learning block and the graph learning block. The temporal convolutional layer and the kernel-level attentive fusion layer are shown in the temporal learning block (A). The local and global graph-filtering layers are shown in the graph learning block (B). The temporal convolutional layer aims to learn dynamic temporal representations from EEG directly instead of human extracted features. The kernel-level attentive fusion layer will fuse the information learned by different temporal kernels to increase the learning capacity of LGGNet. The local graph-filtering layer learns the brain activities within each local region. Then the global graph-filtering layer with a trainable adjacency matrix will be applied to learn complex relations among different local regions. Four local graphs are shown in the figure for illustration purposes only, the detailed local-global-graph definitions are provided in `Defining local-global graphs of EEG' of section III.B. Best viewed in color.}
    \label{fig:LGG}
\end{figure*}
\section{Related Work}
\subsection{Different Representations of EEG Data}
EEG data have two dimensions: channels (EEG electrodes) and time. The channel dimension reflects the brain activities across different functional areas due to different locations of electrodes on the surface of the human's head. The channel refers to the EEG electrodes if not specified. The time dimension contains the changes in brain activities over time. There are three types of EEG representations commonly used in recent studies, namely 2D time-series, images, and graphs. For 2D time-series formats, the network input layer typically consists of temporal convolutional layers to extract temporal information channel by channel and spatial convolutional layers to extract spatial information \cite{doi:10.1002/hbm.23730, Lawhern_2018, 8914184}. Another type of EEG representation is the image. In this, the electrodes are rearranged into a 2D frame based on their relative locations on the brain surface, and the raw data or features of each electrode will be the third dimension of the 2D map \cite{Li2018, 8713896}. Recently, many studies \cite{8320798, 9091308,9309090} have represented EEG data as graphs. In these studies, EEG signals are treated as graphs, with the electrodes being the node and spatial distance or correlations being the edges.
\subsection{Graph Neural Networks}
A graph is represented as $\mathcal{G} = (\mathcal{V}, \mathcal{E})$, where $\mathcal{V}$ is the set of nodes, and $\mathcal{E}$ is the set of edges. $v_i\in \mathcal{V}$ denotes a node, and $e_{i,j}= (v_{i}, v_{j})\in  \mathcal{E}$ denotes an edge.
The adjacency matrix $\textbf{\textit{A}}$ is derived as an n × n matrix with
$\textbf{\textit{A}}_{i,j} = 1$ if $e_{i,j}\in \mathcal{E}$ and $\textbf{\textit{A}}_{i,j}= 0$ if $e_{i,j}\notin \mathcal{E}$. A graph, also known as attributed graph, may have node attributes $\textbf{\textit{X}}$, where $\textbf{\textit{X}} \in\mathbb{R}^{n\times d} $ is a node feature matrix with $\textbf{x}_{v}\in\mathbb{R}^{d}$ representing the feature vector of a node $v$. A graph can be a directed graph or an undirected one. The adjacency matrix of a directed graph may not be asymmetric if a single-direction connection exists. (e.g., $e_{i,j} \neq e_{j,i}$). The adjacency matrix of an undirected graph is symmetric, and $\textbf{\textit{A}} = \textbf{\textit{A}}^{T}$.
GNN \cite{4700287} was proposed to deal with the graph-structured data. Graph Convolutional Neural Networks (GCNN) \cite{NIPS2016_6081} extended the convolution operation to graph in the spectral domain. It can generate a node representation by aggregating its features and neighbors’ features. Kipf \textit{et al.} \cite{DBLP:journals/corr/KipfW16} proposed a scalable graph convolution neural network, which can encode both local graph structure and the feature of the node with improved computational efficiency. 

\subsection{Graph Neural Networks for EEG}
We review some related works that use GNNs to decode EEG signals. Jang \textit{et al.} \cite{8462207} defined the connections as both spatial locations and correlations among electrodes to do video classification via EEG graphs in 2018. In the same year, Song \textit{et al.} \cite{8320798} designed dynamical graph convolutional neural networks for EEG emotion recognition with a trainable adjacency matrix. Lian \textit{et al.} \cite{Lian_2020} refined the graph topology by incorporating the dynamically learned connection weights based on attention and gating mechanisms. GCB-Net \cite{8815811} also utilized a trainable adjacency matrix, and the broad learning system was further applied to learn shallow and deep features. Zhong \textit{et al.} \cite{9091308} defined the adjacency matrix according to the spatial distance and added some global connections according to asymmetry in neuronal activities. GraphNet \cite{9361688} utilized GCN with a distance-based adjacency matrix to decode mental attention states. Instead of learning from hand-crafted features, AMCNN-DGCN  \cite{9309090} learned from EEG directly using multi-scale CNN kernels. After that, GCN layers with a trainable adjacency matrix were applied to learn the spatial relations among electrodes. Although many GNNs were proposed for EEG decoding, most of them didn't model the brain activities within and among different functional areas. 
\begin{table}[!htbp]
\caption{Notations used in Section III}
\begin{center}
\begin{tabularx}{0.45\textwidth} { 
  p{0.1\textwidth}p{0.9\textwidth}
  }
 \hline
\textbf{Symbol}& \textbf{Description}\\
\hline
$f_{S}$& sampling rate\\
$\alpha$& ratio coefficient of temporal kernel size\\
$k, K$& index and number of temporal kernel levels\\
$S_{T}^{k}$& temporal kernel size\\
$\textit{\textbf{X}}_{i}$& EEG samples\\
$i$& index of EEG samples\\
$c$& number of EEG channels\\
$n$& total number of EEG samples\\
$l$& length of the EEG sample in time dimension\\
$\Phi(\cdot)$& activation functions\\
$\textit{\textbf{Z}}$& output tensor of a neural network layer\\
$t$& number of temporal (T) kernels\\
$f$& length of features\\
$\mathcal{F}(\cdot)$& operations\\
$\Gamma(\cdot)$& concatenation of tensors\\
$\mathcal{G}_g$,$\mathcal{G}_f$, $\mathcal{G}_h$, & general, frontal, and hemisphere graphs\\
$\textit{\textbf{A}}_{local}$ & local adjacency matrix\\
$\textit{\textbf{W}}$& trainable weight matrix\\
$\textit{\textbf{b}}$& trainable bias vector\\
$\circ$ & Hadamard product\\
$r, R$& index and total number of local graphs\\
$p, P$& index and number of nodes in a local graph\\
$\textit{\textbf{h}}_{local}$& latent representations of local graphs\\
$\textit{\textbf{A}}_{global}$& global adjacency matrix\\
$\cdot$ & dot product\\
$\textit{\textbf{M}}$ & trainable attentive mask of global adjacency matrix\\
$\textit{\textbf{w}}$& trainable weight\\
$\widetilde{\textit{\textbf{D}}}$ & Degree matrix of the adjacency matrix\\
$h$& length of the hidden output of GCN layers\\
$\Upsilon(\cdot)$& flatten operation\\
\hline
\end{tabularx}
\begin{tablenotes}
      \small
      The order of the symbols is the same as their appearance sequence\\
      \item 
    \end{tablenotes}
\label{Tab:notations}
\end{center}
\end{table}

\section{LGGNet for BCI}
In this section, LGGNet is introduced. Table~\ref{Tab:notations} illustrates the notations used in this section. As shown in Fig.~\ref{fig:LGG}, LGGNet has two main functional blocks, a temporal learning block and a graph learning block. The temporal convolutional layer in the temporal learning block aims to learn dynamic temporal/frequency representations from EEG directly instead of manually extracted features with the help of a kernel-level attentive fusion layer. The graph learning block contains two layers, namely the local and global graph-filtering layers. The local graph-filtering layer learns the brain activities within each neurophysiologically meaningful local region, after which the global graph-filtering layer with a similarity-based trainable adjacency matrix will be applied to learn complex relations among different local regions. 

\subsection{Temporal Learning Block}
Temporal learning block has two modules: temporal convolutional layer and kernel-level attentive fusion layer. 
\subsubsection{Temporal convolutional layer}
The multi-scale temporal convolutional layer utilizes parallel multi-scale 1D temporal kernels (T kernels). In order to learn dynamic-frequency representations, the length of the temporal kernels is set in different ratios of the sampling rate $f_{S}$ \cite{9206750}. The ratio coefficient is denoted as $ \alpha^{k} \in \mathbb{R}$, where $k$ is the level of the temporal convolutional layer. $k$ will vary from 1 to $K$ ($\alpha=0.5, K=3$, in our study). Hence, the size of T kernels in $k$-th level, denoted by $S_{T}^{k}$, can be defined as:
\begin{equation}\label{eq:size_t}
    S_{T}^{k} = \left ( 1, \alpha^{k} \cdot f_{S}\right), k \in [1, 2, 3].
\end{equation}
% The mental state specific information of EEG signals can be reflected in distinct frequency bands \cite{4634130}. Multi-scale 1D temporal convolutional kernels can enrich the learned dynamic time-frequency representations of EEG \cite{9206750}.

Given the preprocessed EEG data $ \textit{\textbf{X}}_{i}\in \mathbb{R}^{c \times l}, i \in \left[ 1,\cdots,n \right ]$, where $n$ equals the number of EEG samples, $c$ is the EEG channel number, and $l$ is the sample length in the time dimension, three multi-scale temporal kernels are applied parallelly to learn dynamic temporal/frequency representations. Instead of using $\Phi_{ReLU}(\cdot)$ as TSception \cite{9206750}, we use the logarithmic of the average pooled square of the representations as \cite{doi:10.1002/hbm.23730} to learn the power features, which are well-studied EEG features in the BCI domain. The 1D CNN layer serves as digital filters which can output filtered signals in different frequency bands \cite{Lawhern_2018}. By squaring the filtered signals, we can get the power of them. An average pooling layer acts as a window function to calculate the averaged power in shorter segments. Then a logarithmic activation is applied as \cite{doi:10.1002/hbm.23730} which shows adding the logarithmic activation can help to improve the performance. Let $\textit{\textbf{Z}}_{temporal}^{k} \in \mathbb{R}^{ t \times c \times f_{k} }$ denote the output of the $k$-th level temporal kernel, where $t$ is the number of the T kernels, and $f_{k}$ is the feature length. $\textit{\textbf{Z}}_{temporal}^{k}$ is defined as:
 \begin{equation}\label{eq:out_T}
     \textit{\textbf{Z}}_{temporal}^{k} = \Phi_{log}(\mathcal{F}_{AP}(\Phi_{square}(\mathcal{F}_{Conv1D}(\textit{\textbf{X}}_{i}, S_{T}^{k})))),
 \end{equation}
 where $\mathcal{F}_{Conv1D}(\textit{\textbf{X}}_{i}, S_{T}^{k})$ is the convolution operation using T kernel of size $S_{T}^{k}$ on $\textit{\textbf{X}}_{i}$, $\Phi_{square}(\cdot)$ is the square function, $\mathcal{F}_{AP}(\cdot)$ is the average pooling operation, and $\Phi_{log}(\cdot)$ is the logarithmic function. The pooling size and step of the $\mathcal{F}_{AP}(\cdot)$ in this power layer were (1, 128) and (1, 0.25*128 = 32) for the attention and the fatigue dataset, the pooling size for DEAP was set as (1, 16) since DEAP had more data that needed a deeper model to learn.
 
 The output of all levels' T kernels will be concatenated along the feature dimension. Hence, the output of the multi-scale temporal convolutional layer for $\textit{\textbf{X}}_{i}$, $\textit{\textbf{Z}}_{MS}^{i} \in \mathbb{R}^{ t \times c \times \sum f_{k} }$, can be calculated by:
  \begin{equation}\label{eq:out_T_multi_scale}
     \textit{\textbf{Z}}^{i}_{MS} = \Gamma(\textit{\textbf{Z}}_{temporal}^{1},\cdots,\textit{\textbf{Z}}_{temporal}^{K}),
 \end{equation}
 where $\Gamma(\cdot)$ is the concatenation operation along the feature (f) dimension. 
 
\subsubsection{Kernel-level attentive fusion}
After concatenation of the output from different level T kernels, a one-by-one convolutional layer is adopted as a kernel-level attentive fusion layer to fuse the features learned by different kernels. Batch normalization \cite{pmlr-v37-ioffe15} is utilized before and after the one-by-one convolution to reduce the internal covariate shift effects. The number of one-by-one kernels is set as $t$. Leaky ReLU is utilized as the activation function. After that, an average pooling layer is utilized to downsample the learned representations. After batch normalization, the fused representations from different one-by-one kernels are then flattened for each EEG channel as its node attribute in EEG-graph representation that will be introduced in the next section. This reshaping process is shown in Fig.~\ref{fig:LGG} (b). Hence, the attentively fused temporal representation of each $\textit{\textbf{X}}_{i}$, $\Bar{\textit{\textbf{Z}}}_{fuse}$, is calculated by:
\begin{equation}\label{eq:final_out_T_fuse}
     \Bar{\textit{\textbf{Z}}}_{fuse}^{i}=\mathcal{F}_{bn}(\mathcal{F}_{AP}(\Phi_{L-ReLU}(\mathcal{F}_{fuse}(\mathcal{F}_{bn}(\textit{\textbf{Z}}_{MS}^{i}))))),
\end{equation}
where $\mathcal{F}_{bn}(\cdot)$ is the batch normalization function, $\mathcal{F}_{fuse}(\cdot)$ is the one-by-one convolution function, and the $\Phi_{L-ReLU}(\cdot)$ is the Leakey ReLU() activation function. The kernel and step sizes of $\mathcal{F}_{AP}(\cdot)$ are both (1, 2). 

The $\Bar{\textit{\textbf{Z}}}_{fuse}^{i} \in \mathbb{R}^{c \times t \times 0.5*\sum f_{k}}$ is reshaped to $\textit{\textbf{Z}}_{fuse}^{i} \in \mathbb{R}^{c \times t *0.5*\sum f_{k}}$ to build the attribute of each node (EEG channel) in the EEG-graph representations:
 
\begin{equation}\label{eq:T_fuse_reshape}
     \textit{\textbf{Z}}_{fuse}^{i} = \mathcal{F}_{reshape}(\Bar{\textit{\textbf{Z}}}_{fuse}^{i}).
\end{equation}

\subsection{Graph Learning Block}
\begin{figure*}[htp]
\centering
    \subfigure[General]{
    \includegraphics[width=0.25\textwidth]{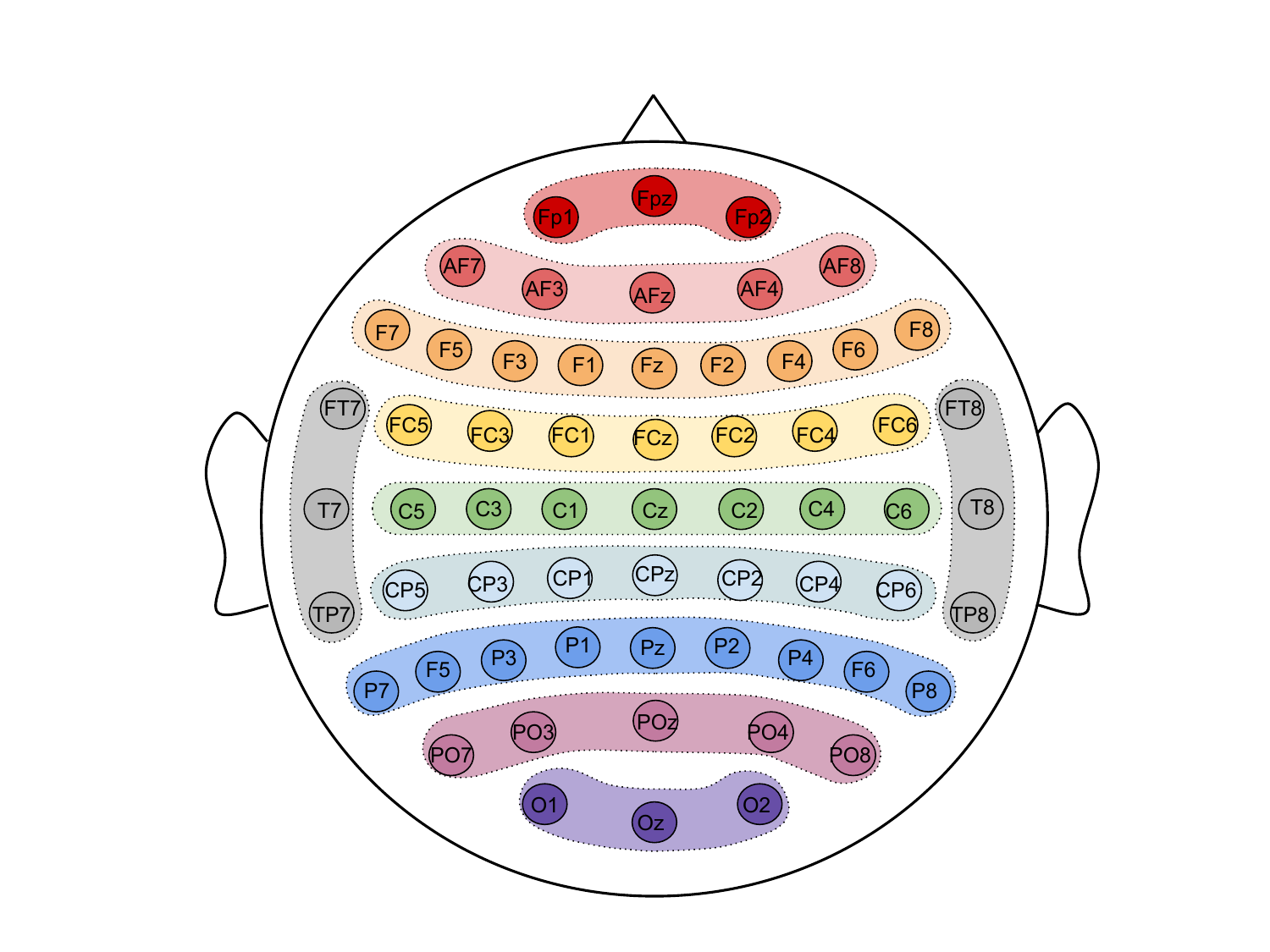}
    }
    \subfigure[Frontal]{
    \includegraphics[width=0.25\textwidth]{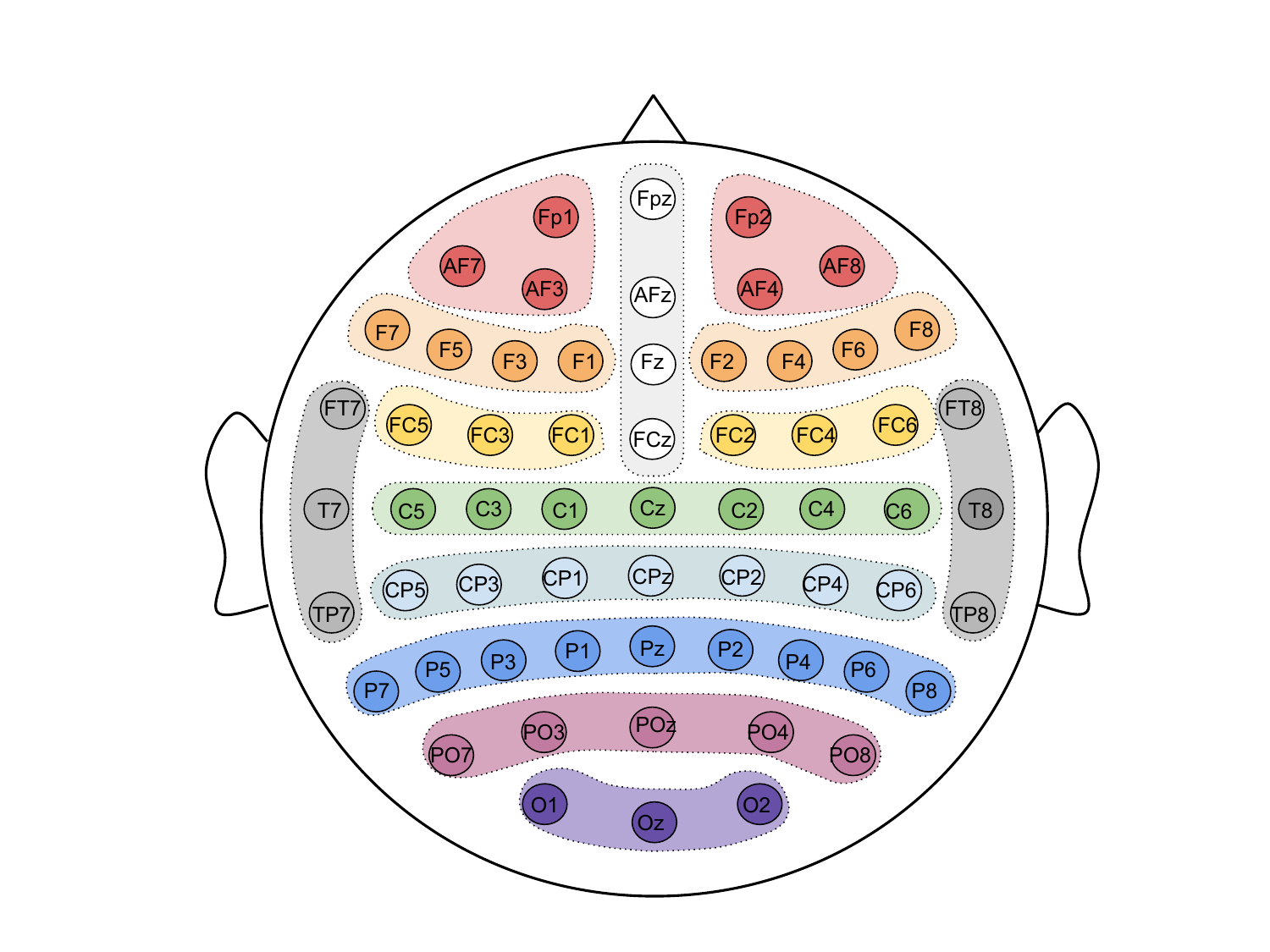}
    }
     \subfigure[Hemisphere]{
    \includegraphics[width=0.25\textwidth]{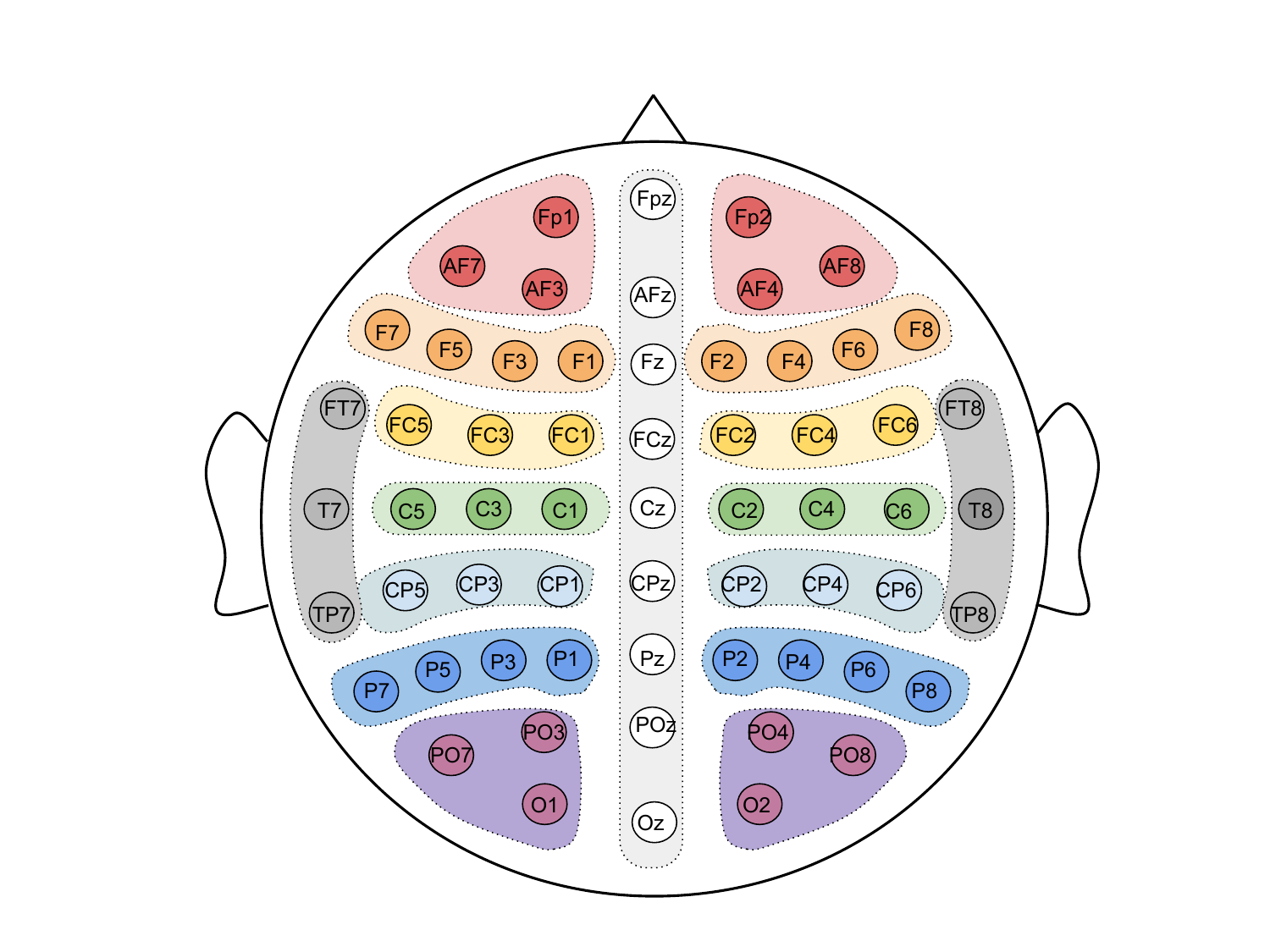}
    }
\caption{Three types of local-global-graph definitions. (a) The general local-global-graph definition. This local graph structure is defined according to the 10-20 system. Each local graph reflects the brain activities of a certain brain functional area. (b) The frontal local-global-graph definition. Based on the general local-global graph, the neuroscience evidence of frontal asymmetry patterns in frontal areas is further considered. Six frontal local graphs that are symmetrically located on the left and right frontal areas of the brain are added to learn more discriminative information. (c) The hemisphere local-global-graph definition. The symmetrical local graphs are added for all the functional areas defined in the general local-global graph. The nodes in a local graph are in the same color. The dotted lines are the local graphs. This diagram illustrates the definition for the 62 channel EEG.}
\label{fig:local_global_graphs}
\end{figure*}
\subsubsection{Defining local-global graphs of EEG}
In this section, three types of local-global-graph representations are constructed based on neuroscience findings \cite{POWER2011665, KOBER2008998, doi:10.1111/psyp.13028}, namely general local-global graph $\mathcal{G}_{g}$, frontal local-global graph $\mathcal{G}_{f}$, and hemisphere local-global graph $\mathcal{G}_{h}$. Given $\textit{\textbf{Z}}_{fuse} \in \mathbb{R}^{c \times t *0.5*\sum f_{k}}$, each electrode is regarded as one node in the EEG graph, and the learned dynamic temporal representations of electrodes are regarded as the node attributes. To learn more information on graph data, the adjacent relations among nodes are very important. To effectively define adjacent relations, several neuroscience findings are taken into consideration. 

Firstly, we define a general local-global graph. Human brains have several functional regions which will be active during different cognitive processes \cite{POWER2011665}. EEG electrodes are placed on scalp surfaces according to the 10-20 system \cite{7946165} that groups channels according to the location on different functional areas of the brain. We define general local-global-graph, $\mathcal{G}_g$, based on the different functional areas of the brain according to the 10-20 system. It is shown in Fig.~\ref{fig:local_global_graphs} (a). LGGNet using the general local-global-graph definition is regarded as LGGNet-G and may be used for more generalized BCI classification tasks.

The frontal local-global graph is further defined based on several neuroscience findings on cognition and emotion studies. The frontal lobe is responsible for high-level behaviours, such as thinking, attention, and emotions \cite{POWER2011665, EGNER2005969}. The frontal asymmetries of EEG appear both on emotional processes \cite{KOBER2008998} and attentional bias to threat \cite{GRIMSHAW201476}. Hence, the frontal area is further split into several smaller local graphs which are symmetrically located on the left and right frontal hemispheres to learn more discriminative information. The frontal local-global graph, $\mathcal{G}_f$ is shown in Fig.~\ref{fig:local_global_graphs} (b).

For the hemisphere local-global graph, we adopt the definition in \cite{10.3389/fpsyg.2012.00428}, which has symmetrical sub-graphs on the left and right hemispheres for all the functional areas. The hemisphere local-global graph, $\mathcal{G}_h$ is shown in Fig.~\ref{fig:local_global_graphs} (c).

We reorder the EEG channels according to the above local-global graphs. The channels within each local graph are next to each other so that the aggregation operation can be applied in the local graph-filtering layer.

\begin{equation}\label{eq:reorder}
    \textit{\textbf{Z}}_{reorder}^{i} = \mathcal{F}_{reorder}(\textit{\textbf{Z}}_{fuse}^{i}).
\end{equation}

\subsubsection{Local graph-filtering layer}
In order to learn the local brain activities, a local graph-filtering layer is proposed to attentively aggregate the learned representations within each local graph, $\textit{\textbf{Z}}_{reorder} \in \mathbb{R}^{c \times t *0.5* \sum f_{k}}$. In this section, the local connections are defined. Then the local graph-filtering layer is introduced.

The electrodes within one local graph are fully connected. The brain consists of local circuits and functional areas \cite{POWER2011665}. Salvador's study \cite{salvador2005neurophysiological} indicates that the strength of connections among brain regions decays as the physical distance increases. Hence, we hypothesize that different electrodes within a sub-group can reflect the similar brain activities of the corresponding functional areas. The local adjacency matrix $\textit{\textbf{A}}_{local}$ is defined as:
\begin{equation}
    \textit{\textbf{A}}_{local} =\begin{bmatrix} 1 & \cdots & 1 &\\ \vdots & \ddots &\vdots &\\1  & \cdots &1 \end{bmatrix},
\label{adj-local}
\end{equation}
where all elements are 1. The size of the $\textit{\textbf{A}}_{local}$ depends on how many channels are within the local graph. 

There are two steps in the local graph-filtering layer: Local graph filtering and local representation aggregating. Given the trainable local graph-filtering matrix $\textit{\textbf{W}}_{local} \in \mathbb{R}^{c \times t*0.5* \sum f_{k}}$, and local graph-filtering bias vector $\textit{\textbf{b}}_{local} \in \mathbb{R}^{c \times 1}$, the local graph-filtering weights will be assigned to the representation of each electrode by:
\begin{equation}\label{eq:local-weight-assign}
     \textit{\textbf{Z}}_{filtered}^{i} =\Phi_{ReLU}( \textit{\textbf{W}}_{local}\circ\textit{\textbf{Z}}^{i}_{reorder} - \textit{\textbf{b}}_{local}),
 \end{equation}
 where $\circ$ is the Hadamard product.

After local graph filtering, the attentively filtered representation within each local graph will be aggregated by an aggregating function $\mathcal{F}_{aggregate}(\cdot)$ to get the hidden embeddings of the local graphs. 
Let $\textit{\textbf{Z}}_{filtered}^{i} =[{\textit{\textbf{Z}}_{1}}',\cdots,{\textit{\textbf{Z}}_{r}}', \cdots, {\textit{\textbf{Z}}_{R}}']^{T}$ be the locally filtered graph representations, where ${\textit{\textbf{Z}}_{r}}' \in \mathbb{R}^{P_{r} \times {f}'}$ is the local-graph representation, $R$ is the total number of local graphs, $P_{r}$ is the number of nodes in the $r$-th local graph ($\sum{P_{r}}=c$), and ${f}'$ is the feature length of each node after local graph filtering. A local graph can be denoted as: ${\textit{\textbf{Z}}_{r}}'=[\textit{\textbf{z}}^{1}_{r}, \cdots,\textit{\textbf{z}}^{p}_{r}, \cdots, \textit{\textbf{z}}^{P_{r}}_{r}]^{T}$, where $\textit{\textbf{z}}^{p}_{r}$ is the node vector in the local graph. The aggregating function aggregates the node vectors within each local graph. It can be maximum, minimum, average, etc. In LGGNet, the average operation is selected as the aggregating function. Hence, the output of the local graph-filtering layer, $\textit{\textbf{Z}}_{local}^{i} \in \mathbb{R}^{R \times {f}'}$, can be calculated by:
\begin{gather}\label{eq:local-aggregate}
    \nonumber\textit{\textbf{Z}}_{local}^{i} = \mathcal{F}_{aggregate}(\textit{\textbf{Z}}^{i}_{filtered})\\
    \nonumber=\mathcal{F}_{aggregate}([{\textit{\textbf{Z}}_{1}}',\cdots,{\textit{\textbf{Z}}_{R}}']^{T})\\
    =[\frac{1}{P_{1}}\sum^{P_{1}}_{p=1}\textit{\textbf{z}}_{1}^{p}, \cdots, \frac{1}{P_{R}}\sum^{P_{R}}_{p=1}\textit{\textbf{z}}_{R}^{p}]^{T}
    \\
    \nonumber=[\textit{\textbf{h}}_{local}^{1}, ..., \textit{\textbf{h}}_{local}^{R}]^{T},
  \end{gather}
where $r$ is the index of local graphs, $p$ is the index of nodes in each local graph, and $\textit{\textbf{h}}_{local}$ is latent representations of local graphs.

\subsubsection{Global graph-filtering layer}
The graph convolution on the global graph is designed to learn the complex relations among local graphs. We firstly define the global connections after which the details about the global graph-filtering layer are presented in this part. 

The global connection is defined based on the relations among local graphs. Neuroscience studies suggested that activating one particular brain region also tends to activate other regions in the group for the high-level cognitive process \cite{KOBER2008998}. The relations among local graphs are utilized as the edges in the global graph. The dot products between local graph representations for each EEG instance are calculated to reflect the relations among local graphs. Note that, the similarity adjacency matrix is dynamic and instance-specific. We assume the global connection is undirected because the relation between two local graphs is mutual. The basic adjacency matrix of the global graph, $\textit{\textbf{A}}_{global-base} \in \mathbb{R}^{R \times R}$, is symmetric and can be defined as:
\begin{equation}
    \textit{\textbf{A}}_{global-base} =\left.\begin{bmatrix} \textit{\textbf{h}}_{local}^{1} \cdot \textit{\textbf{h}}_{local}^{1} & \cdots & \textit{\textbf{h}}_{local}^{1} \cdot \textit{\textbf{h}}_{local}^{R}&\\ \vdots & \ddots &\vdots &\\\textit{\textbf{h}}_{local}^{1} \cdot \textit{\textbf{h}}_{local}^{R} & \cdots &\textit{\textbf{h}}_{local}^{R} \cdot \textit{\textbf{h}}_{local}^{R}& \end{bmatrix}\right.\label{adj-global},
\end{equation}
where $\cdot$ is the dot product. 

Due to the complex relations among brain functional areas, a trainable attentive mask is adapted to emphasize the most important connections in the instance-level similarity adjacency matrix. Note that the trainable mask is also symmetric because the global adjacency matrix is undirected. The trainable attentive mask, $\textit{\textbf{M}} \in \mathbb{R}^{R \times R}$, can be defined as:
\begin{equation}
    \textit{\textbf{M}} =\left.\begin{bmatrix} \textit{\textbf{w}}_{1,1} & \cdots & \textit{\textbf{w}}_{1,R}&\\ \vdots & \ddots &\vdots &\\\textit{\textbf{w}}_{R,1} & \cdots &\textit{\textbf{w}}_{R,R}& \end{bmatrix}\right.\label{adj-global-weight},
\end{equation}
where $\textit{\textbf{w}}$ are the trainable parameters, and $\textit{\textbf{w}}_{1,R}=\textit{\textbf{w}}_{R,1}$. 

Self-loops are added after applying the trainable mask to the basic global adjacency matrix. Because adding self-loops after applying the trainable mask can maintain the strength of the self-loops since the values in the trainable mask are generally small. The ReLU activation function is applied to make the adjacency matrix non-negative. Hence, the final global adjacency matrix can be calculated as:
\begin{equation}\label{eq:final_adj_global}
    \textit{\textbf{A}}_{global} = \Phi_{ReLU}(\textit{\textbf{A}}_{global-base} \circ \textit{\textbf{M}}) + \textit{\textbf{I}},
\end{equation}
where $\textit{\textbf{I}} \in \mathbb{R}^{R \times R}$ is the identity matrix.

% Having the node attributes, $\textit{\textbf{Z}}_{fuse} \in \mathbb{R}^{c \times t *0.5*\sum f_{k}} $, and the adjacency matrix of the local-global graphs, graph filtering are designed to learn the local-global graph representations in the next two sections. 

Given the global adjacency matrix,  $\textit{\textbf{A}}_{global} \in \mathbb{R}^{R \times R}$ described in \textbf{Eq}.\ref{eq:final_adj_global}, a GCN \cite{DBLP:journals/corr/KipfW16} layer is adopted to learn the global-graph representations. The normalized adjacency matrix, $\widetilde{\textit{\textbf{A}}}_{global}$, can be calculated by:
\begin{equation}\label{eq:norm-adj}
     \widetilde{\textit{\textbf{A}}}_{global}=\widetilde{\textit{\textbf{D}}}^{-\frac{1}{2}}\textit{\textbf{A}}_{global}\widetilde{\textit{\textbf{D}}}^{-\frac{1}{2}},
 \end{equation}
where $\widetilde{\textit{\textbf{D}}}=\sum_{q}\textit{\textbf{A}}^{p,q}_{global}$ is the degree matrix of the $\textit{\textbf{A}}_{global}$.

Before the global graph filtering, batch normalization, $\mathcal{F}_{bn}(\cdot)$, is applied. In LGGNet, the number of global GCN layers is set to be one. Let the projecting weight matrix of GCN layer be $\textit{\textbf{W}}_{global} \in \mathbb{R}^{{f}' \times h}$, where $h$ is the length of the hidden output after GCN, and the trainable bias vector be $\textit{\textbf{b}}_{global} \in \mathbb{R}^{h \times 1}$. The global-graph filtering of $\textit{\textbf{Z}}_{local}^{i}$ can be calculated by:
\begin{equation}\label{eq:global-filtering}
     \textit{\textbf{Z}}_{global}^{i} = 
     \Phi_{ReLU}( \widetilde{\textit{\textbf{A}}}_{global}(\mathcal{F}_{bn}(\textit{\textbf{Z}}_{local}^{i})\textit{\textbf{W}}_{global} - \textit{\textbf{b}}_{global} )).
 \end{equation}
After getting the globally filtered representation, batch normalization is applied. Then the flattened representation will be fed into a linear layer to generate the final classification output as:
\begin{equation}\label{eq:fc}
     Output = \Phi_{softmax}(\textit{\textbf{W}}\mathcal{F}_{dropout}(\Upsilon(\mathcal{F}_{bn}(\textit{\textbf{Z}}_{global}^{i}))) + \textbf{\textit{b}}),
 \end{equation}
 where the $\Upsilon(\cdot)$ is the flatten operation, $\textit{\textbf{W}}$ is the trainable weight matrix, and $\textbf{\textit{b}} \in \mathbb{R}^{n_{classes} \times 1}$ is the bias term ($n_{classes}$ is the number of classes which is two in this paper).
 
%needed in second column of first page if using \IEEEpubid
%\IEEEpubidadjcol
The structure of graph learning block is summarized in Table \ref{Tab:strc_table}. It shows the operations, input, and output of each module. A single output sample whose size is $(c \times {f}')$ from the temporal learning block is used for the illustration. 

\begin{table*}[ht]
  \caption{Structure of Local and Global graph filtering.}
\makebox[\textwidth][c]{
\begin{tabular}{c|c|c|c}
\toprule
Layer  & Operations & Input & Output \\
\midrule
Local-graph filtering&Filtering&$Z_{reorder}(c \times {f}')$ & $Z_{filtered}(c \times {f}')$\\ \cline{2-4}
&Aggregation& $Z_{filtered}(c \times {f}')$ and graph definitions &$Z_{local}(R \times {f}')$\\
\hline
Global-graph filtering&Graph convolution&$Z_{local}(R \times {f}')$ and $A_{global}(R \times R)$ & $Z_{global}(R \times h)$\\ \hline
Output layer &MLP and Softmax & $Z_{global}(R \times h)$&$Output(n_{classes})$\\
\bottomrule
\end{tabular}
}
\begin{tablenotes}
      \small
      \item The ${f}'= t*0.5* \sum f_{k}$ is the feature length of each node after the temporal learning block.
    \end{tablenotes}
\label{Tab:strc_table}
\end{table*}

 Finally, the proposed LGGNet can be summarized in Algorithm~\ref{alg:LGG}

\begin{algorithm}[t]
\SetAlgoLined
\SetKwBlock{DoParallel}{do in parallel}{end}
\SetKwBlock{DoSequential}{do in sequential}{end}
\KwIn{ EEG data $\textit{\textbf{X}}_{i} \in \mathbb{R}^{c \times l}$; ground truth label $y$; graph definitions $\mathcal{G}_{g}$, $\mathcal{G}_{f}$, and $\mathcal{G}_{h}$; global adjacency matrix $\textit{\textbf{A}}_{global}$ }

\KwOutput{$pred$, the prediction of LGGNet}
Initialization\;
%\com{\# \textit{get the output of the temporal convolutional layer}} \
\For{$j\gets1$ \KwTo $3$}{
            get $j$-th temporal kernel size by \textbf{Eq.~\ref{eq:size_t}}\;
            get $\textbf{Z}_{temporal}^{j}$ by \textbf{Eq.~\ref{eq:out_T}} using $\textit{\textbf{X}}_{i}$ as input\;
        }
get $\textbf{Z}_{T-MS}^{i}$ by \textbf{Eq.~\ref{eq:out_T_multi_scale}}\;
do kernel-level attention fusion by \textbf{Eq.~\ref{eq:final_out_T_fuse}} to get $\textbf{Z}_{fuse}^{i}$\;
%\com{\# \textit{get the output of the local graph-filtering layer}} \\
do local filtering on each node attribute by \textbf{Eq.~\ref{eq:local-weight-assign}}\;
aggregate the filtered node attribute within each local graph ($\mathcal{G}_{g}$, $\mathcal{G}_{f}$, or $\mathcal{G}_{h}$) by \textbf{Eq. 9}\;
%\com{\# \textit{get the output of the global graph-filtering layer}} \\
get the $\textit{\textbf{A}}_{global}$ by \textbf{Eq.~\ref{adj-global} -~\ref{eq:norm-adj}}\;
do global filtering on embeddings of local graphs by \textbf{Eq.~\ref{eq:global-filtering}} with $\widetilde{\textit{\textbf{A}}}_{global}$\;
get $pred$ by \textbf{Eq.~\ref{eq:fc}}\;
\KwReturn{$pred$}
\caption{LGG}
\label{alg:LGG}
\end{algorithm}

%\textbf{LGG-G(ours)} &\textbf{64.53}(12.37)& \textbf{64.40}(14.84)& 89.62(4.47)& \textbf{87.63}(9.55)& \textbf{59.19}(7.82)& \textbf{64.51}(10.37)&62.86(13.44) & 72.42(13.21)\\
\section{Experiments}
\subsection{Datasets}
Three publicly available datasets of different tasks were utilized to evaluate the proposed LGGNet: the attention dataset \cite{Shin2018} for attention classification, the fatigue dataset \cite{cao2019multi} for fatigue classification, and the DEAP dataset \cite{5871728} for emotion and preference classification, respectively. 

The attention dataset\footnote{http://doc.ml.tu-berlin.de/simultaneous\_EEG\_NIRS/} is a multimodal brain-imaging dataset to measure three cognitive tasks of healthy subjects. The discrimination/selection response task (DSR) was involved in this paper for cognitive attention classification. 26 subjects participated in the experiment. The first session among the three was utilized for each subject to avoid the effects of cross-session variance. There were several series of attention task periods (40s) and rest periods (20s) in each session. 28 EEG channels and 2 electrooculography (EOG) channels were recorded with a sampling rate of 1K Hz. 

The fatigue dataset\footnote{https://figshare.com/articles/dataset/MultichannelEEGrecordingsduringasus\\tained-attentiondrivingtask/6427334} provides the EEG signals to measure the cognitive fatigue states of the driver during a 90-min-long driving task in a VR driving environment. 27 subjects participated in the data collection experiments. The subjects were introduced to keep the car cruising in the center of the lane while random lane-departure events were induced. 32-channel EEG signals were collected with a sampling rate of 500Hz. 

DEAP\footnote{http://www.eecs.qmul.ac.uk/mmv/datasets/deap/index.html} is a multi-modal human affective states dataset, including EEG, facial expressions, and galvanic skin response (GSR). 40 1-minute-long emotional music videos were used to induce different emotions to the subject. Before each trial, there was a 3-second baseline. Subjects provided their self-assessments on arousal, valence, dominance, and liking after each trial, using a continuous 9-point scale. The valence and liking dimensions were utilized for the emotion and preference classification tasks in this paper. 32 subjects participated in the data collection experiments. 32-channel EEG signals were recorded with a sampling rate of 512 Hz. 

\subsection{Pre-processing}
EEG signals with several pre-processing operations were used as the input samples of the neural networks instead of hand-crafted features.

For the attention dataset, a band-pass filter from 0.5-50 Hz was applied to remove low and high-frequency noise as \cite{9361688}. EOG was removed using the automatic ICA EOG removal method in the MNE toolbox \cite{GRAMFORT2014446}. Then the data were downsampled to 200 Hz. Following \cite{ZHANG2021129}, only the first half of each attention trial was utilized to balance the samples between attention and inattention (rest). Each trial was further segmented into 4-second segments with a 50\% overlap.

For the fatigue dataset, the officially preprocessed EEG dataset \cite{cao2019multi} was used in this paper. The raw EEG signals were band-passed from 1 to 50 Hz. Eye blinks were removed by visual checking. The Automatic Artifact Removal (AAR) method in EEGLab \cite{DELORME20049} was used to remove ocular and muscular artifacts. The processed data were downsampled to 128 Hz as \cite{8918989}. For fatigue level calculation, we also followed \cite{8918989}. The 3s' EEG data before the onset of the lane-departure events were used as EEG trials. Reaction time (RT) was utilized to measure the fatigue level for the EEG trials. RT was defined as the time from the onset of the lane-departure event to the onset of the counter-steering event. The RT of one trial was defined as local RT, denoted by $RT_{l}$ . The global RT ($RT_{g}$) of the one trial was the mean of the local RTs of all the trials within a 90-second window before the current trial. The 5th percentile of all local RTs in the entire session was selected as an alert RT, $RT_{a}$. Let 0 be the label of the fatigue class, and 1 be the non-fatigue class, the labeling process can be defined as:
\begin{equation}\label{eq:fatigue-labeling}
     y = \left\{\begin{matrix}
0 & RT_{l} > 2.5*RT_{a} \&\& RT_g > 2.5*RT_{a}\\ 
1 & RT_{l} < 1.5*RT_{a} \&\& RT_g < 1.5*RT_{a}
\end{matrix}\right.
 \end{equation}
We followed \cite{8918989}, only the subjects whose number of the smaller class trial was larger than 50 was utilized for evaluation. However, we didn't balance the data as \cite{8918989} did, so that more data was available to train the network and our proposed method was able to classify unbalanced data.

For DEAP, the processed data provided by the author was utilized. First, the 3 seconds' pre-trial baseline was removed from each trial. After that, the data were down-sampled to 128 Hz. EOG was removed using the method described in \cite{5871728}. A band-pass filter from 4 to 45 Hz was applied. Then the average reference was conducted on the filtered data. To divide each dimension into high/low classes, 5 was chosen as the threshold to project the continuous 9-point scale into low and high classes in each dimension as \cite{5871728, 9141493}. Each trial was further split into 4s shorter non-overlapping segments to train the neural network. 

% \subsection{Evaluation Metrics}
% The accuracy and F1 score were used as the evaluation metrics. The reason for using accuracy is that it is one of the most commonly used evaluation metrics in classification problems. F1 score was also utilized. Because it can provide a more comprehensive evaluation of the classifier when the dataset has unbalanced classes. For the fatigue and DEAP datasets, the classes are unbalanced. Hence, using both accuracy and F1 score can provide a better evaluation of the classifier. 

% The calculating formulas of accuracy and F1 score are illustrated as follows: 
% \begin{equation}\label{eq:acc}
%      Accuracy = \frac{TP+TN}{TP+FP+TN+FN}
%  \end{equation}

% \begin{equation}\label{eq:f1}
%      F1 = \frac{TP}{TP+\frac{1}{2}(FP+FN)}
%  \end{equation}
%  where $TP$ is the true positive, $TN$ is the true negative, and $FP$ is the false positive, and $FN$ is the false negative.

\begin{table*}[ht]
\caption{Comparison between our proposed LGGNet against SOTA classifiers on three benchmark datasets using trial-wise n-fold cross-validation.}
\makebox[1\linewidth][c]{
\begin{tabular}{l|ll|ll|ll|ll}
\toprule
 &\multicolumn{2}{c|}{Attention}& \multicolumn{2}{c|}{Fatigue} & \multicolumn{2}{c|}{Emotion}&
 \multicolumn{2}{c}{Preference} \\
\midrule
%model &mean ACC& mean F1 & mean ACC& mean F1& mean ACC& mean F1&mean ACC& mean F1\\ \hline
model &mACC (\%)& mF1 (\%) & mACC (\%)& mF1 (\%)& mACC (\%)& mF1 (\%)&mACC (\%)& mF1 (\%)\\ \hline
DeepConvNet \cite{doi:10.1002/hbm.23730} & 58.97* & 64.30& 66.86**& 74.79**&58.67 & 57.48*&61.98 &70.59 \\
EEGNet \cite{Lawhern_2018}& 58.05*& 59.06 &87.62 & 84.85& 56.38& 60.03** &58.41** & 67.45***\\
TSception \cite{9206750}& 57.76*&57.93* &86.17** & 84.77*& 57.46**& 60.42***& 61.70*& 70.34***\\
R2G-STNN \cite{8736804}& 57.76* & 57.99*& 88.79**& 88.42*& \textbf{60.11}& 63.40&60.99** & 69.89**\\
HRNN \cite{9361688}& 56.84* &55.20* &77.96*& 72.76& 58.46& 60.93* & 62.74& 71.31*\\ 
RGNN \cite{9091308}& 55.48*** &55.86** &83.55**& 79.91*& 57.90& 61.28& \textbf{63.60}& \textbf{73.90}\\
AMCNN-DGCN \cite{9309090}& 51.25*** &51.71** &78.95***& 75.72**& 52.91***& 55.51***& 61.95& 70.94\\
GraphNet \cite{9361688}& 55.41** &57.92* &78.78***& 74.48*& 53.83***& 54.59***& 62.06& 71.79\\ 

\hline
\textbf{LGGNet-H} &61.22& 60.08& 89.83& 89.14&58.85 & 64.15&\textbf{63.07} &\textbf{72.53}\\
\textbf{LGGNet-F} &63.07& 60.63& \textbf{90.76}& \textbf{90.18}& 58.80& 63.68& 62.86 & 71.96\\
\textbf{LGGNet-G} &\textbf{64.53}& \textbf{64.40}& 90.14& 89.31& \textbf{59.19}& \textbf{64.51}&62.86 & 72.42\\
\bottomrule
\end{tabular}%
}
\begin{tablenotes}
      \small
      \item $p$-value of the improvement of LGGNet over the method: * indicating $(p < 0.05)$, ** indicating $(p<0.01)$, *** indicating $(p < 0.001)$.
     
      \item LGGNet-H, LGGNet-F, LGGNet-G: LGGNet using hemisphere, frontal, and general local-global graphs.
    \end{tablenotes}
\label{Tab:ACC-F1}
\end{table*}%

\subsection{Experiment Settings}
Trial-wise n-fold cross-validation for subject-specific experiments was adopted to evaluate the proposed LGGNet. In subject-specific experiments, the training and test data are all from the same subject. 

To avoid potential data leakage issues caused by improper random shuffling in subject-specific experiments, we adopted trial-wise shuffling instead of segment-wise shuffling. For the continuous cognitive processes in the brain, such as attention and emotion, the adjacent data segments in one trial are highly correlated. If one randomly shuffles the segments before the training-testing split, the highly correlated segments will appear in both training and test data. Hence, a very high classification result will be observed. However, the accuracy will drop when the highly correlated segments are never seen by the model in a real-world situation. For the attention and DEAP datasets, each trial was split into shorter segments as \cite{9206750}. The trial-wise shuffling ensures that the highly correlated segments within a trial do not appear in both train and test data in a cross-validation fold.

The nested cross-validation \cite{Varma2006} was utilized to avoid biased evaluation. The outer loop of the nested cross-validation was the trial-wise n-fold cross-validation, and the inner loop was another k-fold cross-validation, where $n_{DEAP} = n_{fatigue} = 10$, $n_{attention} = 6$, and $k = 3$ in this work. The mean accuracy and F1 score of all subjects were reported as the final evaluation criterion as \cite{5871728}. In the inner loop, to make full use of the training data, a two-stage training strategy was utilized as well. More details about the two-stage training process are provided in the next section.

\subsection{Two-stage Training}
The optimization process via two-stage training is introduced here. To make full use of the training data, for each step of trial-wise n-fold cross-validation, the neural networks were trained in two stages using the training data. Since the inner loop of the nested cross-validation was the k-fold cross-validation, one fold of training data was utilized as validation data in each step of the k-fold cross-validation. First, the best-performing model in the k folds was saved as the candidate for testing. Then all k folds of the training data were combined as the new training data. The candidate model was fine-tuned on the combined training data with a smaller learning rate compared with the first stage training. In the second stage, the pre-trained model was trained for a maximum of 20 epochs. The training process stopped when the training accuracy reached 100\% the first time to make sure the model was well fine-tuned without over-fitting. Test data was not used in any step of the two-stage model training. After getting the fine-tuned model, it was evaluated on the test data.

\subsection{Implementation Details}
The code was implemented using PyTorch \cite{NIPS2019_9015} library, and the source code can be found via this link\footnote{https://github.com/yi-ding-cs/LGG}.

Cross-entropy loss was selected as the objective function to guide the training process. For model training, the maximum training epoch of the first stage was 200 while the one for stage II was 20 instead. The batch size was 64. The dropout rate was set as 0.5 for all three datasets. Adam optimizer was utilized to optimize the training process with the initial learning rate being 1e-3 which was scaled down by a factor of 10 in the second stage. For the attention dataset, we used 1e-2 as the initial learning rate because it yielded higher validation accuracy. Early stopping was applied to reduce the training time and overcome over-fitting. We set the hidden size of GCN to 32 and the number of T kernels to 64 for all three datasets. We tuned the pooling size of the power layer on the attention dataset based on the performance on the validation set and applied the same value to the fatigue dataset. Note the hyper-parameter settings were the same for all the subjects within each dataset. Label smoothing with a 0.1 smoothing rate was applied when training networks on DEAP dataset because the classes were highly unbalanced for some subjects. For more details, please refer to the open-access GitHub repository for LGGNet.

\section{Results and Discussion}
% To evaluate the proposed LGGNet, we tested it on three publicly available datasets for cognitive attention \cite{Shin2018}, mental fatigue \cite{cao2019multi}, emotion, and preference \cite{5871728} four different types of cognitive tasks. The accuracy and F1 score were used as the evaluation metrics. 

The performances of LGGNet were compared with CNN, RNN, and GNN-based SOTA methods in the BCI domain. The CNN-based methods include: \textbf{DeepConvNet} \cite{doi:10.1002/hbm.23730}, \textbf{EEGNet} \cite{Lawhern_2018}, and \textbf{TSception} \cite{9206750}. The RNN-based methods include: \textbf{R2G-STNN} \cite{8736804} and \textbf{HRNN} \cite{9361688}. The GNN-based methods include: \textbf{RGNN} \cite{9091308}, \textbf{AMCNN-DGCN} \cite{9309090}, and \textbf{GraphNet} \cite{9361688}. All the methods were under the same generalized evaluation settings which were utilized to avoid data leakage issue. For fair comparisons, all the baseline methods used the optimal parameters suggested by their authors and we used the same training codes and settings as that of LGGNet. 

In this section, We first show the accuracies and F1 scores against the SOTA methods with statistical analysis. Extensive analysis experiments were conducted to understand LGGNet better, including ablation studies, the effect of the local-global graphs, and the effect of the activation function in temporal convolutional layer. Then saliency maps were utilized to visualize the most informative region of the data identified by LGGNet. The learned adjacency matrices were visualized to see what relations of the local graphs were learned by LGGNet.

\subsection{Statistical Analysis}
We first report the mean accuracy (ACC) and mean F1 score on the three benchmark datasets for four types of cognitive tasks (shown in Table~\ref{Tab:ACC-F1}). The two-tailed Wilcoxon Signed-Rank Test was utilized for the statistical analysis on the attention dataset and DEAP, while paired T-test was used on the fatigue dataset because there were fewer subjects in the fatigue dataset.

\subsubsection{Attention Classification Task}
LGGNet-G achieves the highest classification results in most of the experiments, especially for the attention dataset, on which the improvements in accuracies are all statistically significant. The accuracies of LGGNet-G are 9.12\% ($p<0.01$), 13.28\% ($p<0.001$), 9.05\% ($p<0.001$), 7.69\% ($p<0.05$), 6.77\% ($p<0.05$), 6.77\% ($p<0.05$), 6.48\% ($p<0.05$), and 5.56\% ($p<0.05$) higher than these of GraphNet, AMCNN-DGCN, RGNN, HRNN, R2G-STNN, TSception, EEGNet, and DeepConvNet, respectively. The improvements achieved by LGGNet-G in F1 scores over these baselines are 6.48\% ($p<0.05$), 12.69\% ($p<0.01$), 8.54\% ($p<0.01$), 9.20\% ($p<0.05$), 6.41\% ($p<0.05$), 6.47\% ($p<0.05$), 5.34\% ($p=0.091$), and 0.10\% ($p=0.928$) respectively. 

\subsubsection{Fatigue Classification Task}
On the fatigue dataset, the best accuracy and F1 score are achieved by LGGNet-F with most of the improvements being statistically significant. LGGNet-F achieves 90.76\% ACC in fatigue detection tasks, which are 11.98\% ($p<0.001$), 11.81\% ($p<0.001$), 7.21\% ($p<0.01$), 12.8\% ($p<0.05$), 1.97\% ($p<0.01$), 4.59\% ($p<0.01$), 3.14\% ($p=0.231$), and 23.9\% ($p<0.01$) higher than the ones of GraphNet, AMCNN-DGCN, RGNN, HRNN, R2G-STNN, TSception, EEGNet, and DeepConvNet, respectively. The improvements in F1 scores over these baselines are 15.7\% ($p,0.05$), 14.46\% ($p<0.01$), 10.27\% ($p<0.05$), 17.42\% ($p=0.068$), 1.76\% ($p<0.05$), 5.41\% ($p<0.05$), 5.33\% ($p=0.302$) and 15.39\% ($p<0.05$), respectively. 

\subsubsection{Emotion Classification Task}
LGGNet-G still achieves the highest F1 score (64.51\%) in the emotion classification task, while the best accuracy is achieved by R2G-STNN (60.11\%). The differences in accuracies on the DEAP dataset are less than the ones on the other datasets, but all the LGGNet variants achieve relatively larger improvements over the baselines. Compared with GNN-based methods, LGGNet-G has 5.36\% ($p<0.001$), 6.28\% ($p<0.001$) and 1.29\% ($p=0.242$) higher ACC than GraphNet, AMCNN-DGCN, and RGNN. And the improvements in F1 scores are 9.92\% ($p<0.001$), 9.00\% ($p<0.001$), and 3.23\% ($p=0.126$). LGGNet-G achieves higher ACCs and F1 scores than all the RNN and CNN-based baselines, except R2G-STNN. 

\subsubsection{Preference Classification Task}
LGGNet-H achieves the highest accuracy (63.07\%) and F1 score (72.53\%) in the preference classification task among three variants of LGGNet, while RGNN achieves the highest ACC (63.60\%) and F1 score (73.90\%) among all the compared methods. But the performance differences between LGGNet-H and LGGNet-G are not significant. Except for RGNN, LGGNet-H has higher ACCs and F1 scores than the compared baseline methods. Especially for GNN-based methods, LGGNet-H has 1.01\% ($p=0.271$) and 1.12\% ($p=0.190$) higher ACCs than GraphNet and AMCNN-DGCN. And LGGNet-H has 0.74\% ($p=0.358$) and 1.59\% ($p=0.052$) higher F1 scores than GraphNet and AMCNN-DGCN. 

\begin{table}[t]
  \caption{Results of ablation studies on DEAP using LGGNet-H.}
\makebox[0.5\textwidth][c]{
\begin{tabular}{ccc|c|c|c|c}
\toprule
AF & L  &  G  & ACC(\%) & Changes(\%)& F1(\%) & Changes(\%)\\
\midrule
&\cmark& \cmark&60.95 &\drop{-2.12} &68.45 &\drop{-4.08}\\ 
\cmark& &\cmark&60.93 &\drop{-2.14} &69.36 &\drop{-3.17}\\ 
\cmark& \cmark& & 59.06& \drop{-4.01}& 67.08&\drop{-5.45}\\\hline

\cmark&\cmark&\cmark &\textbf{63.07} &- &\textbf{72.53} &-\\
\bottomrule
\end{tabular}%
}
\begin{tablenotes}
      \small
      \item\cmark: Keep the component.
      \item  AF: Kernel-level attentive fusion. L: Local graph-filtering layer. G: Global graph-filtering layer.
      \item Changes: Compared with the original LGGNet-H.
    \end{tablenotes}
\label{Tab:ablation}
\end{table}%

\begin{table}[t]
  \caption{Effect of the activation function in the temporal convolutional layer of LGGNet-H on DEAP.}
\makebox[0.5\textwidth][c]{
\begin{tabular}{c|c|c|c|c}
\toprule
Activation function  & ACC(\%)& changes(\%) & F1(\%) & changes(\%)\\
\midrule
Power$\rightarrow$Leaky-ReLU&61.64 & \drop{-1.43}&70.35 &\drop{-2.18}\\
Power$\rightarrow$ELU&61.21 & \drop{-1.86}&69.71 & \drop{-2.82}\\
Power$\rightarrow$SELU& 60.98&\drop{-2.09} & 69.39&\drop{-3.14}\\ \hline
Power &\textbf{63.07} &- &\textbf{72.53} &-\\
\bottomrule
\end{tabular}
}
\begin{tablenotes}
      \small
      \item Changes: Compared with the original LGGNet-H
    \end{tablenotes}
\label{Tab:effect_act_fun}
\end{table}

\subsection{Ablation Study}
To better understand the individual contribution of the components kernel-level attentive fusion, local graph filtering, and global graph filtering in LGGNet, ablation studies were conducted by removing each of these blocks from the LGGNet-H network. DEAP dataset is utilized because there are more data and subjects compared to the other datasets. The ablation studies were conducted on the preference classification task because the performances were better than the ones for the emotion classification task using DEAP. Hence, LGGNet-H was utilized because it achieved the best classification results among the proposed methods. In the first ablation study, to investigate the contribution of the kernel-level attentive fusion, this block was removed from the network and the output of the temporal convolutional layer was reshaped as node by feature and was sent directly to the local-graph filtering layer. In the second study, the learned temporal representations from the temporal learning block were used as the input of the global-graph filtering layer directly to obtain performance without local graph filtering. Finally, the feature output from the local-graph filtering layer was flattened and passed to the MLP to get the output of network without global graph filtering. The new classification accuracies and the performance changes are reflected in Table~\ref{Tab:ablation}.

\subsubsection{The Contribution of The Kernel-level Attentive Fusion}
According to results shown in the first row of Table~\ref{Tab:ablation}, removing the kernel-level attentive fusion makes the accuracy drop from 63.07\% to 60.95\%, decreasing by 2.12\%. For the F1 score, it even drops more with the decrease being 4.08\%. The results show the effectiveness of the kernel-level attentive fusion. 

\subsubsection{The Contribution of The Local Graph Filtering}
To understand the contribution of the local graph-filtering layer, it was removed from LGGNet. In this case, each EEG channel is one node in the graph and the global adjacency matrix, $\textbf{\textit{A}}_{global} \in \mathbb{R}^{c \times c}$, reflects the connection among all the nodes ($c$ is the number of EEG electrodes). 

According to the results shown in the second row of Table~\ref{Tab:ablation}, after removing the local graph-filtering layer entirely, the accuracy drops from 63.07\% to 60.93\%, decreasing by 2.14\%. For the F1 score, it drops by 3.17\%. This indicates the importance of the local graph-filtering layer.

\subsubsection{The Contribution of The Global Graph Filtering}

The global graph filtering was removed from the LGGNet to analyze its importance to the classification performance. In this situation, only the local graph-filtering layer was kept to learn the spatial pattern of EEG. After getting the embeddings of local graphs, the latent representation was fed into fully connected layers without global graph filtering.

According to the third row of Table~\ref{Tab:ablation}, the accuracy and F1 score all dropped after removing the global graph-filtering layer. And the decreases are higher than the ones without the local graph-filtering layer. A 4.01\% drop was observed for accuracy after discarding the global graph-filtering layer, while the one for the F1 score was 5.45\%. The results show the contribution of the global graph filtering is larger than the one of local graph filtering in LGGNet.  

\begin{figure}[tp]
    \centering
    \includegraphics[width=0.8\linewidth]{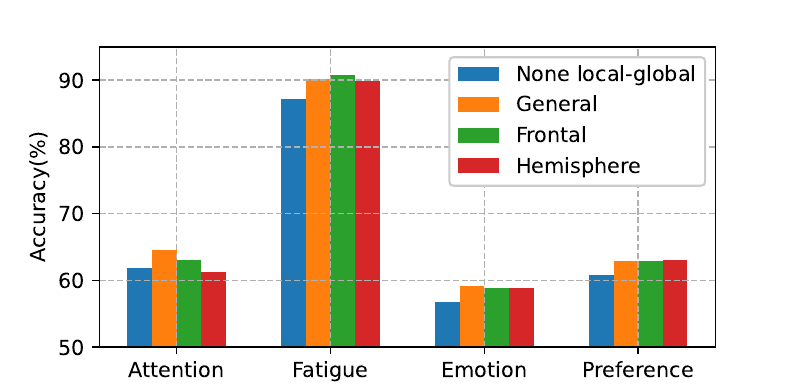}
    \caption{Mean accuracies of LGGNet using different graph structures. The blue bar is the baseline that has no local-global graphs.}
    \label{fig:bar_ACC}
\end{figure}

\begin{figure}[tp]
    \centering
    \includegraphics[width=0.8\linewidth]{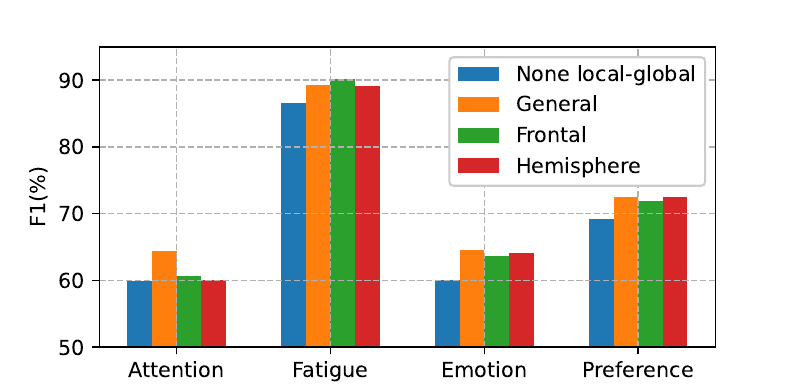}
    \caption{Mean F1 scores of LGGNet using different graph structures. The blue bar is the baseline that has no local-global graphs.}
    \label{fig:bar_F1}
\end{figure}

\subsection{Effects of Activation Functions in The Temporal Convolutional Layer}
To study the effects of different activation functions, we replaced the power layer with commonly used activation functions, such as ReLU, Leaky-ReLU, ELU, and SELU separately. 

Replacing the power layer with other commonly used activation functions causes the decrease of the classification results. The results are shown in Table~\ref{Tab:effect_act_fun}. Using Leaky-ReLU() in the temporal convolutional layer has the least drops in accuracy (1.43\%) and F1 score (2.18\%). The largest drops were observed when we replace the power layer with SELU() activation function, which are 2.09\% in terms of accuracy and 3.14\% in terms of F1 score. This indicates the importance of the power layer.

\subsection{Effects of Local-Global Graphs}
To evaluate the effects of treating EEG as local-global graphs that were specially designed according to neuroscience, we compare them with a none local-global graph baseline. Only global graph convolution was conducted because there were no local graphs in the baseline. The effects of different local-global-graph definitions were also analyzed by comparing their performances on different cognitive tasks. The results are shown in Fig.~\ref{fig:bar_ACC} and ~\ref{fig:bar_F1}. And the detailed accuracies and F1 scores of three LGGNet variants are shown in the last three rows of Table ~\ref{Tab:ACC-F1}. 

Using local-global graphs that are specially designed according to neuroscientific evidence yields significant improvements on classification performances for all four cognitive tasks, except the ones for the attention classification task when frontal and hemisphere local-global graphs were used in LGGNet. Compared with the baseline, LGGNet-G achieves 2.63\% ($p=0.087$) and 4.50\% ($p<0.05$) higher accuracy and F1 score than those of the baseline for the attention classification. For the fatigue detection task, the improvements achieved by using LGGNet-G are 2.99\% ($p<0.01$) and 2.75\% ($p<0.01$). In the emotion classification task, a 2.49\% ($p<0.001$) higher accuracy and a 4.48\% ($p<0.001$) higher F1 score are observed when the general local-global graph is used than the ones of the baseline that uses the none local-global graph. The improvements achieved by LGGNet-G on the preference classification task are 2.26\% ($p<0.01$) and 3.33\% ($p<0.001$) in terms of accuracy and F1 score. The results indicate the effectiveness of using local-global graphs to extract the spatial information of EEG.

The general local-global graph has a higher generalization ability as expected. LGGNet using the $\mathcal{G}_{g}$ achieves the highest classification accuracies and F1 scores for both attention and emotion classification tasks. However, in the mental fatigue classification task, LGGNet-F achieves the highest F1 score and the highest accuracy. LGGNet-H achieves the highest classification results for the preference classification task instead. But the differences in performance are not significant for fatigue, emotion, and preference classification tasks. This suggests adding more symmetric local graphs in functional areas can yield certain improvements over the general local-global graph for some tasks but the improvements are not significant.

\begin{figure*}[htp]
\centering
    \subfigure[Attention]{
    \includegraphics[width=0.22\textwidth]{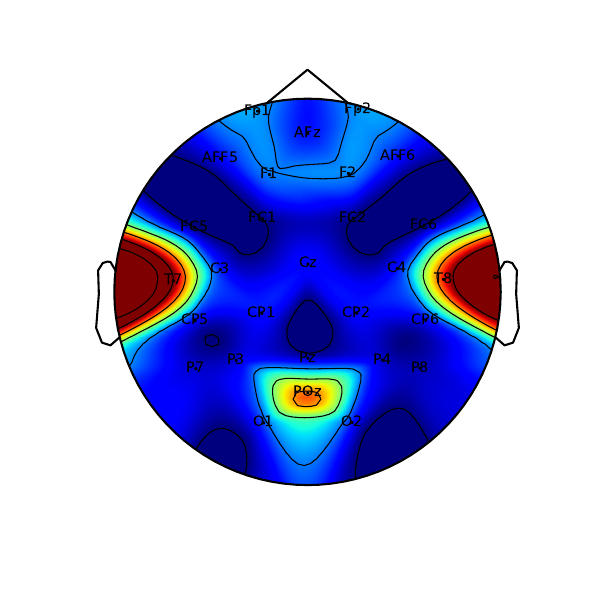}
    }
    \subfigure[Fatigue]{
    \includegraphics[width=0.22\textwidth]{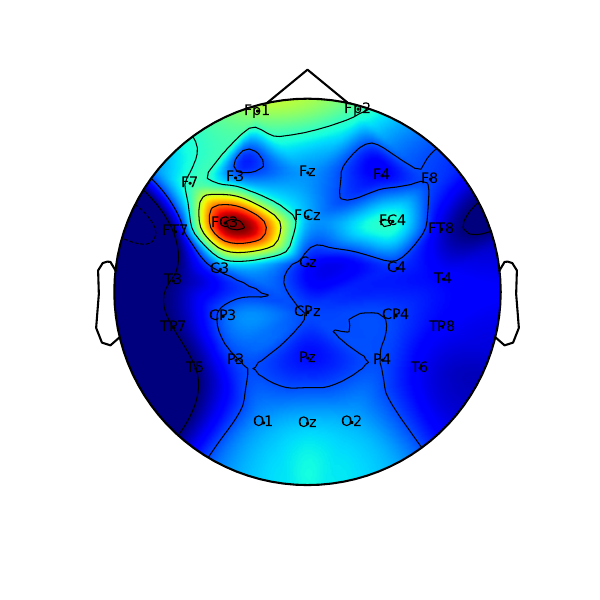}
    }
     \subfigure[Emotion]{
    \includegraphics[width=0.22\textwidth]{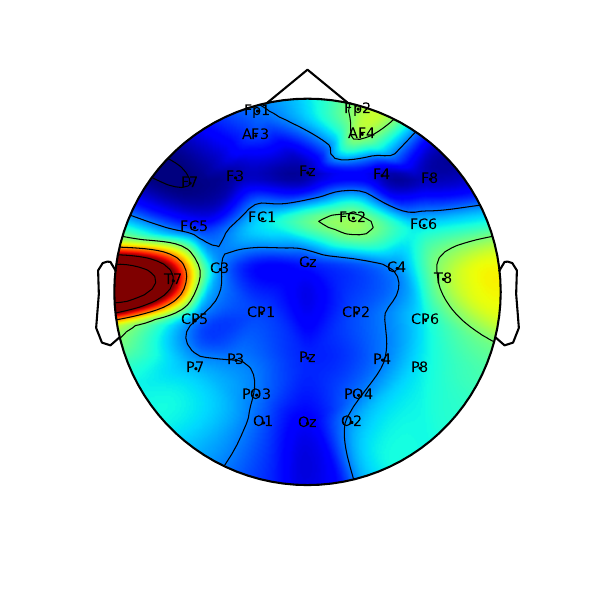}
    }
    \subfigure[Preference]{
    \includegraphics[width=0.22\textwidth]{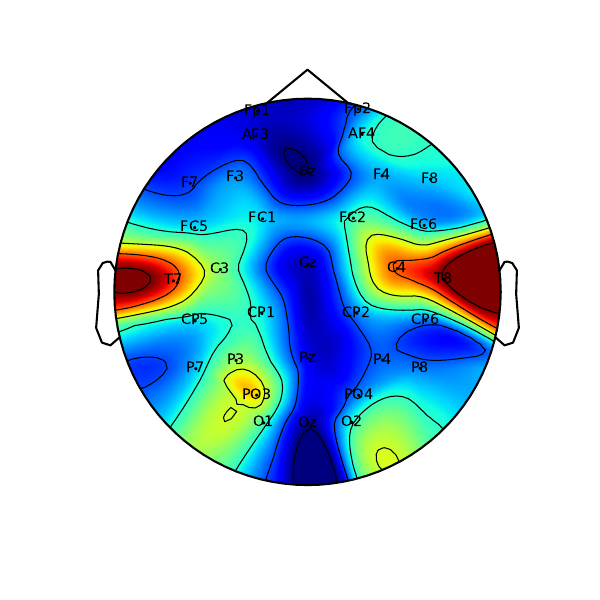}
    }
\caption{Mean saliency maps of all subjects for three datasets. These mean saliency maps are for: (a) attention, (b) fatigue, (c) emotion, and (d) preference. }
\label{fig:saliencymap}
\end{figure*}

\begin{figure*}[htp]
\centering
    \subfigure[Attention]{
    \includegraphics[width=0.22\textwidth]{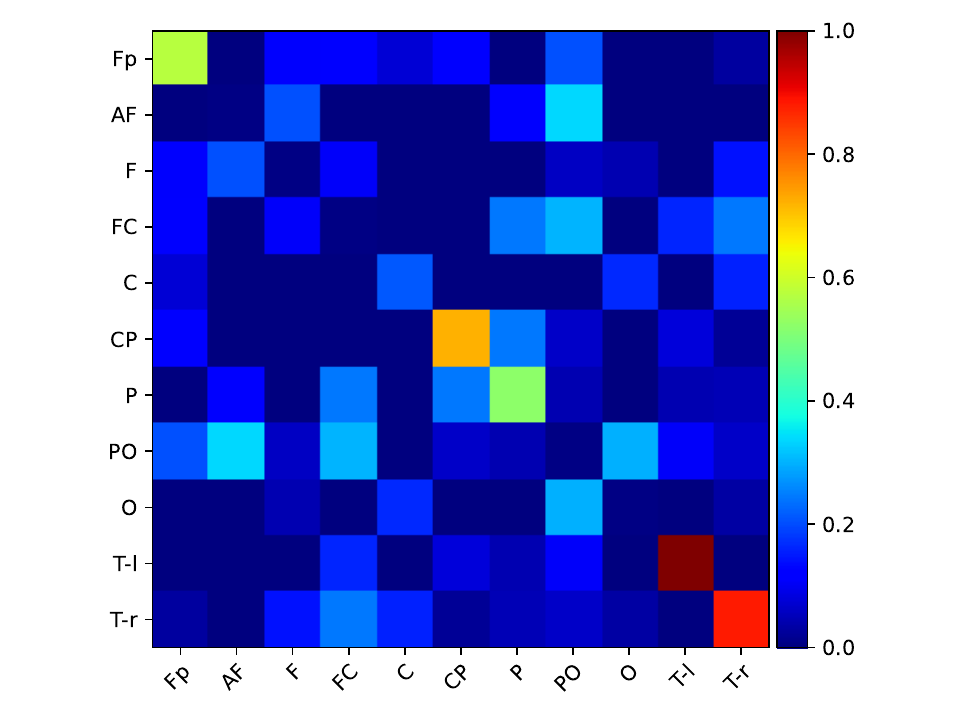}
    }
    \subfigure[Fatigue]{
    \includegraphics[width=0.22\textwidth]{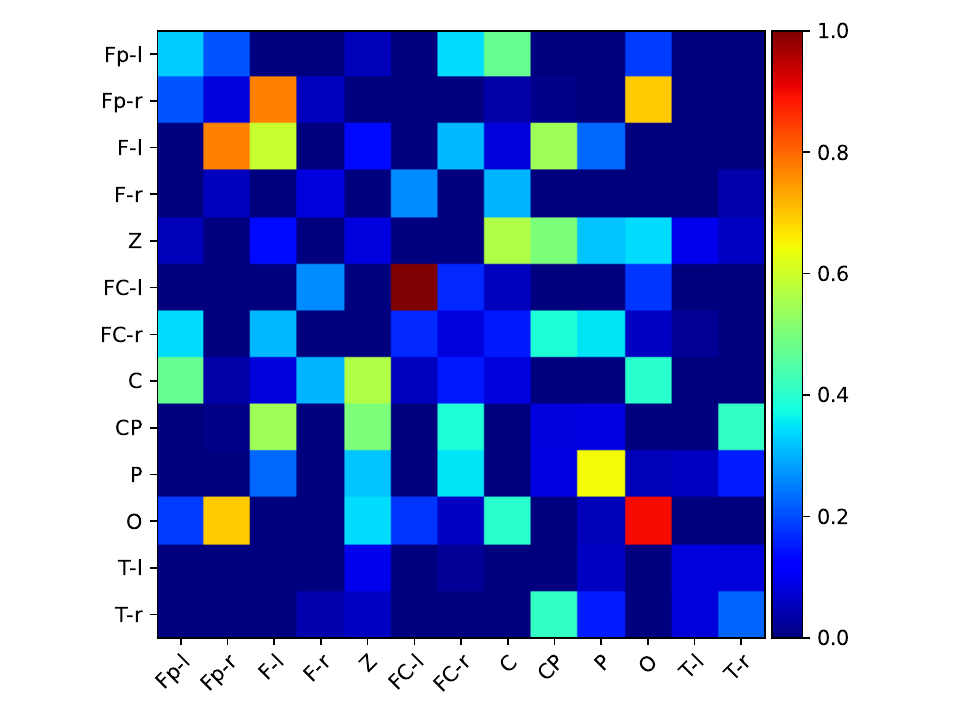}
    }
     \subfigure[Emotion]{
    \includegraphics[width=0.22\textwidth]{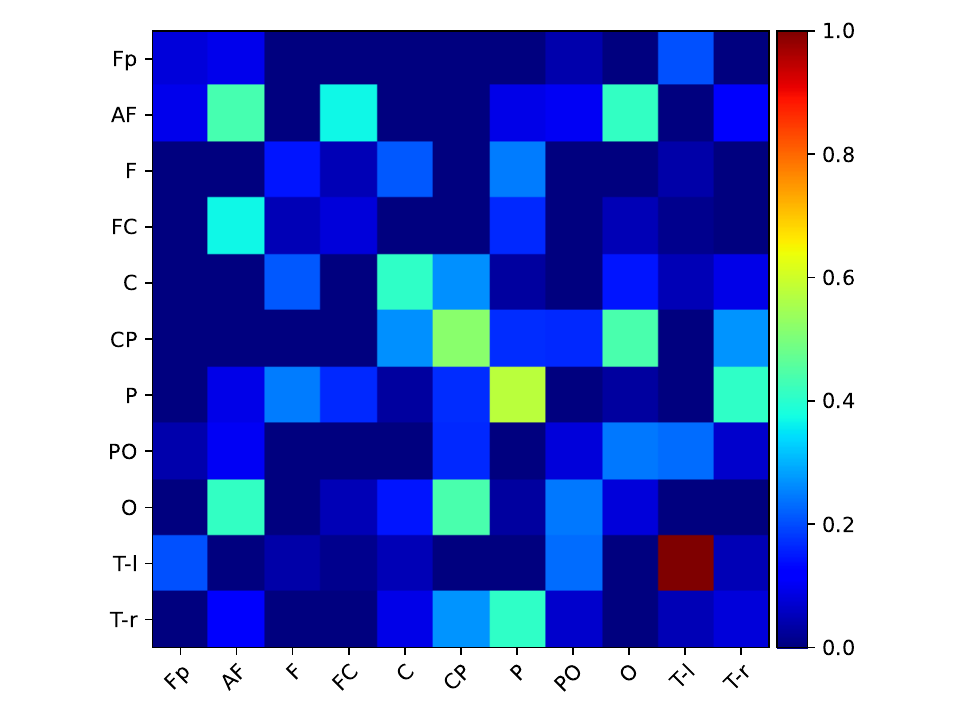}
    }
    \subfigure[Preference]{
    \includegraphics[width=0.22\textwidth]{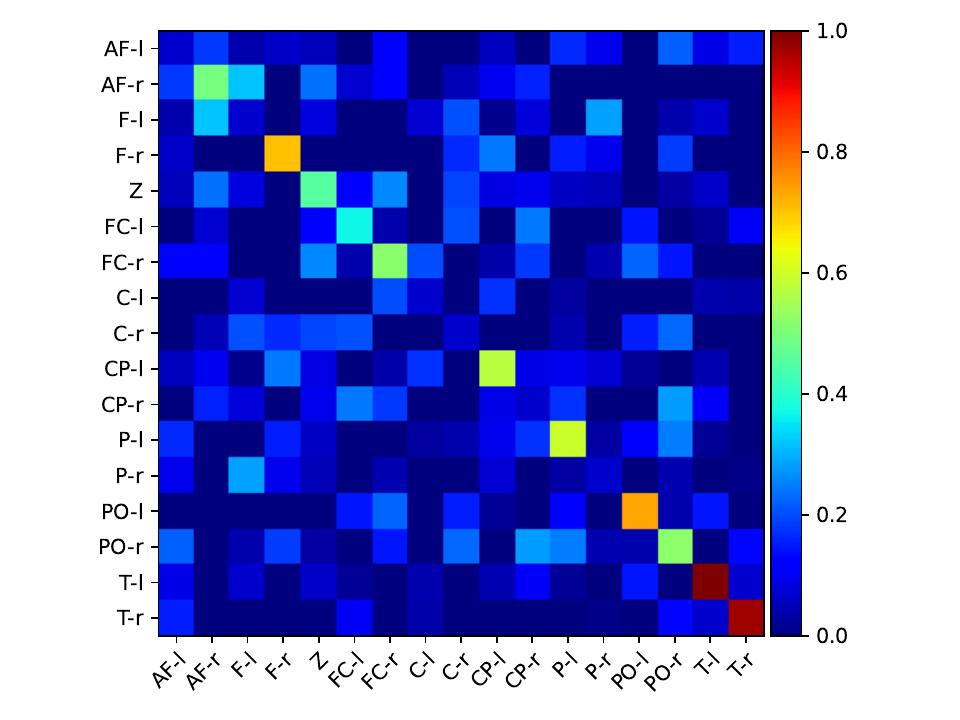}
    }
\caption{Visualization of the final learned adjacency matrices for four cognitive tasks. These mean adjacency matrices are for: (a) attention, (b) fatigue, (c) emotion, and (d) preference.}
\label{fig:adjs}
\end{figure*}

\begin{figure*}[htp]
\centering
    \subfigure[Attention]{
    \includegraphics[width=0.22\textwidth]{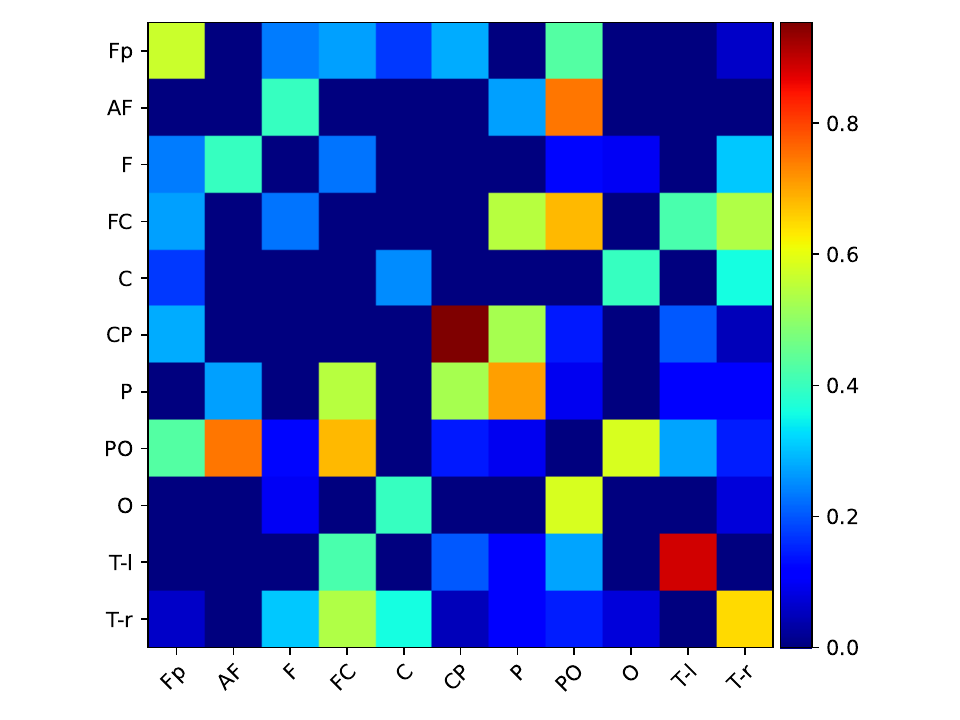}
    }
    \subfigure[Fatigue]{
    \includegraphics[width=0.22\textwidth]{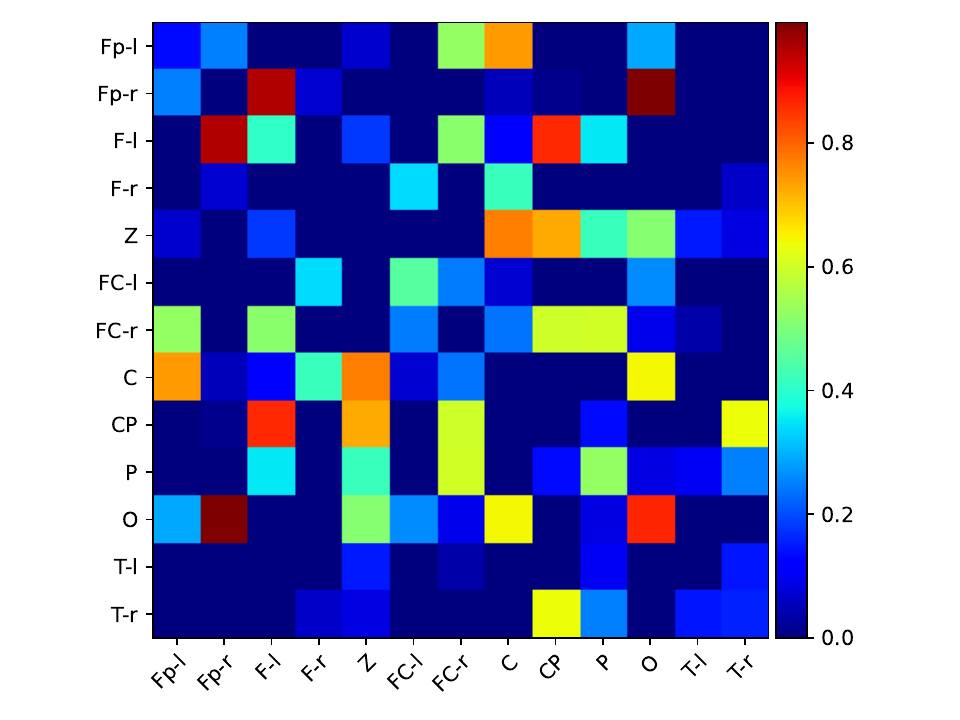}
    }
     \subfigure[Emotion]{
    \includegraphics[width=0.22\textwidth]{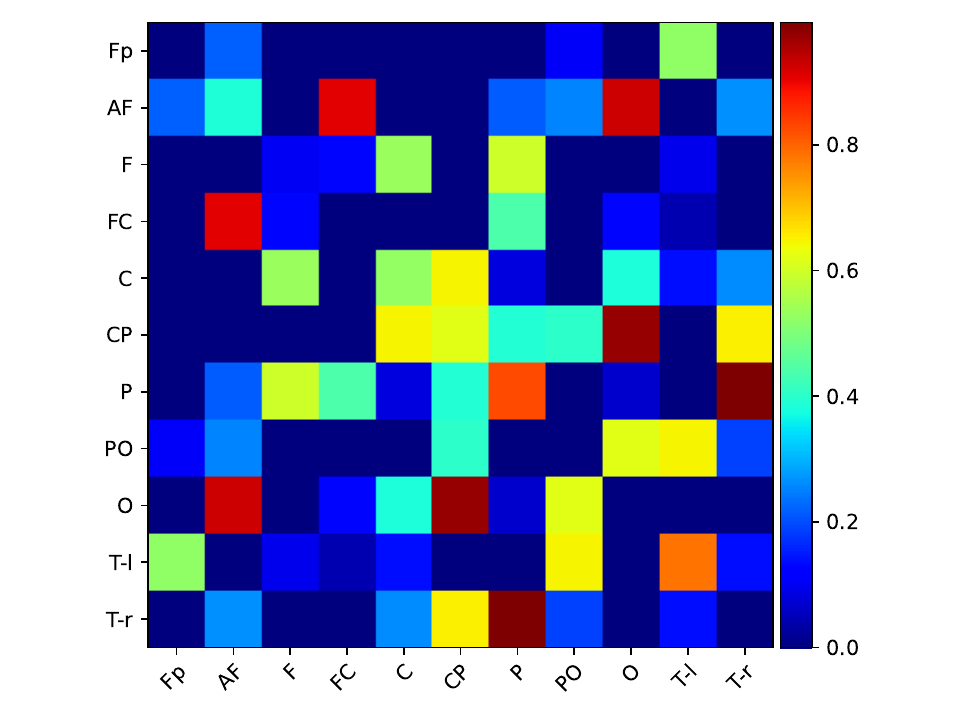}
    }
    \subfigure[Preference]{
    \includegraphics[width=0.22\textwidth]{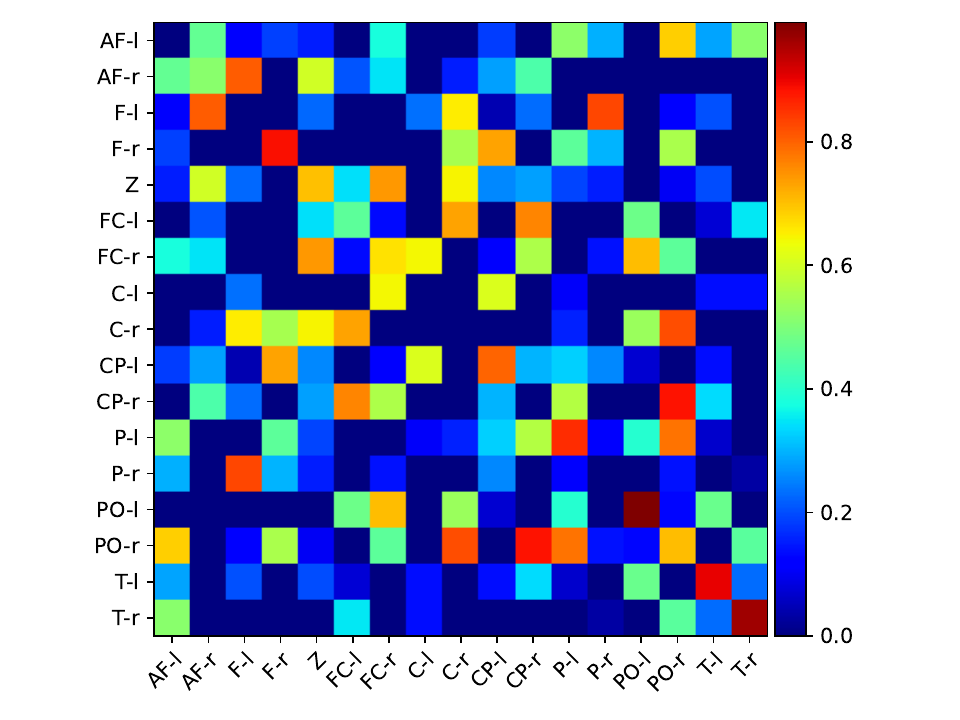}
    }
\caption{Visualization of the learnable attentive masks for four cognitive tasks. These mean attentive masks are for: (1) attention, (b) fatigue, (c) emotion, and (d) preference.}
\label{fig:adjs_mask}
\end{figure*}

% add the analysis according to the ablation study
\subsection{Interpretability and Visualization}
\subsubsection{Saliency Maps Visualization}
In this part, the saliency map \cite{simonyan2013deep} is utilized to visualize which parts of the data are more informative. To better visualize the saliency map, the original saliency map was averaged along the time dimension to get the topological map of the EEG channels for each subject. Fig.~\ref{fig:saliencymap} shows the averaged saliency map of all the subjects.

From Fig.~\ref{fig:saliencymap} (a), LGGNet mainly learns from temporal (T7 and T8), and parietal area (POz) of the brain for attention classification. This is also suggested in \cite{10.3389/fnhum.2018.00222, 10.1371/journal.pone.0050590} that the temporal and parietal lobes are attention-related regions. 

The frontal area provides more fatigue classification-related information to LGGNet. According to Fig.~\ref{fig:saliencymap} (b), strong activations are observed on Fp1, Fp2, F7, FC3 and FC4. This is consistent with other studies \cite{JING2020366} which indicate the frontal lobe is related to human fatigue states. 

From Fig.~\ref{fig:saliencymap} (c), LGGNet learns more emotional information from the temporal (T7 and T8) and frontal (Fp2, AF4, and FC2) areas of the brain. Temporal area, especially the left temporal area provides more emotion related information, which is in line with \cite{olson2007enigmatic}. Frontal area also contributes to the final classification results. This is consistent to \cite{7946165, kragel2018generalizable}, which indicate the frontal area is related to emotions. 

According to Fig.~\ref{fig:saliencymap} (d), LGGNet learns more from the temporal areas (T7 and T8) of the brain. Neuroimaging study \cite{CHAN2019391} suggests the temporal lobes are predictive for the preference prediction during video watching, and these brain areas are related to sensory integration and emotional processing.   

The above neurological knowledge indicates the neural network learns from the task-related regions of EEG signals. 

\subsubsection{Learned Global Connection Visualization}
In this section, the final learned adjacency matrices and learnable attentive masks of each task's best performing model are visualized in Fig.~\ref{fig:adjs} and ~\ref{fig:adjs_mask} to understand what relations LGGNet learns from EEG for different cognitive tasks. To get a general view of each cognitive task, the normalized learnable attentive masks and adjacency matrices are averaged for all the subjects. All the negative values in the learnable attentive masks were set zero before normalization because of the ReLU activation function in equation~\ref{eq:final_adj_global}. Because the adjacent relations are among local graphs instead of individual EEG channels, the names of the local graphs are defined by the name of the functional area. The 'l' and 'r' are utilized to indicate the location of the symmetric sub-graphs within a functional area. 

Attention: Some connections between frontal and parietal regions, between (1) AF and PO and (2) FC and PO, are observed in the adjacency matrix shown in Fig.~\ref{fig:adjs} (a) for attention classification task. And for the self-connections, frontal (Fp), parietal (CP and P), and temporal (T-l and T-r) have higher attentive weights. We further visualize the learned attentive mask to see the learned relations among different local areas. According to Fig.~\ref{fig:adjs_mask} (a), some connections between frontal and parietal are enhanced by the attentive mask. Besides, the self-loops of CP and T-l get more attention weights. It is consistent with \cite{10.1371/journal.pone.0050590} which indicates the posterior parietal lobe (PPL) that has dense connectivity with the cortical and subcortical regions in frontal, temporal, and occipital lobes.

Fatigue: According to Fig.~\ref{fig:adjs} (b), relatively stronger connections are observed among frontal sub-areas (between Fp-r and F-l) for the fatigue classification task. The connections between frontal and occipital areas, frontal and motor areas are also relatively strong. These connections are between (1) Fp-r and O, (2) F-l and CP, (3) Fp-l and C, and (4) C and Z. For the self-connections, more attentive weights are given to frontal (Fp-l, F-l, and FC-l), parietal (P), and occipital (O). The frontal lobe and parietal areas are related to mental attention functions \cite{10.1371/journal.pone.0050590}. The relatively strong connections among frontal, occipital, and motor areas may be because the visual and motor processes were involved in the fatigue experiment (driving in VR). For the learned attentive mask, according to Fig.~\ref{fig:adjs_mask} (b), some connections between frontal and occipital are enhanced by the attentive mask. For the self-loops, the one of area O has the highest attention weight.

Emotion: More connections among frontal, occipital, and temporal, which are between (1) AF and FC, (2) AF and O, (3) CP and O, and (4) P and T-r, are observed in the final learned adjacency matrix (Fig.~\ref{fig:adjs} (c)) than the ones for the attention and preference classification tasks. It is also consistent with neuroscience \cite{KOBER2008998} that the emotional process involves more basic processes, such as perception and attention. For self-connections, the frontal and temporal areas, commonly known as the emotion-related areas, have higher attention weights than the others. However, C, CP, and P also get some attention weights. It may be caused by the attention function involved in the high-level emotional processes. For the attentive mask shown in Fig.~\ref{fig:adjs_mask} (c), the connections that are enhanced by the attentive mask are between (1) FC and AF, (2) O and AF, (3) O and CP, and (4) T-r and P. For the self-loops, the ones of P and T-l get more attention.    

Preference: We find that there are fewer connections among different local graphs in the final learned adjacency matrix of the preference prediction task shown in Fig.~\ref{fig:adjs} (d). More attentive weights are given to the temporal area (T-l, T-r) than the other regions. But the frontal (AF-r, F-r, FC-r) and occipital (PO-l and PO-r) areas are also highlighted. For the attentive mask shown in Fig.~\ref{fig:adjs_mask} (d), the connections that are enhanced by the attentive mask are between (1) F-l and AF-r, (2) P-r and F-l, (3) PO-r and C-r, and (4) PO-r and CP-r. Among self-loops, F-r, CP-l, P-l, PO-l, T-l, and T-r get higher attention weights than the others. 

Visualizing the learned adjacency matrices and the learnable attentive masks shows the relations and the important local regions identified by LGGNet. And most of the learned relations are task-related. Since the cognitive processes are complex and may involve more basic processes that are not unique to the task, more analysis should be conducted in the future to better understand what and how the network learns from EEG.

Although LGG achieves the highest classification results for most of the experiments for four cognitive tasks, the limitation of this work should also be noticed. In this work, the nodes within each local area are set to be fully connected, which might not be able to reflect the complex brain activities inside that functional area. How to model the relations within local areas should be explored. The learned connections in the adjacency matrices of attention, emotion, and preference task are not as strong as the ones of fatigue task. Further improvement of the network or loss function design should be considered in the future to improve the classification performance.

\section{Conclusion}
In this paper, we propose LGGNet, a neurologically inspired graph neural network, to learn from local-global-graph representations of EEG. Multi-scale 1D temporal convolutional kernels with kernel-level attention fusion are utilized to learn the temporal dynamics of EEG. Local and global graph filtering learn the brain activities within each functional area and the complex relations among them during the cognitive process in the brain, respectively. With a robust nested cross-validation strategy, the proposed method and several state-of-the-art methods are evaluated on three publicly available benchmark datasets for attention, fatigue, emotion, and preference classification tasks. The proposed method achieves significantly ($p<0.05$) higher accuracies and F1 scores than other methods in most of the experiments. Further analyses also show that applying neuropsychological knowledge to the network design ensures that networks are trained on task-specific neural activations.

% \section*{Appendix}

\section*{Acknowledgment}
This work was partially supported by the RIE2020 AME Programmatic Fund, Singapore (No. A20G8b0102). The authors would like to thank Ms. Aidi Liu and Mr. Sun Hao for their help on the proofreading of the manuscript.  

\ifCLASSOPTIONcaptionsoff
  \newpage
\fi

\bibliographystyle{./Transactions-Bibliography/IEEEtran}
\bibliography{./Transactions-Bibliography/mybib}

\end{document}